\documentclass{article}

\usepackage{microtype}
\usepackage{graphicx}
\usepackage{subfigure}
\usepackage{booktabs} 

\usepackage[accepted]{icml2024}

\usepackage{xcolor}         
\usepackage[utf8]{inputenc} 
\usepackage[T1]{fontenc}    
\usepackage{url}            
\usepackage{booktabs}       
\usepackage{amsfonts}       
\usepackage{nicefrac}       
\usepackage{microtype}      
\usepackage{xcolor}         
\usepackage{longtable}
\usepackage{booktabs}
\usepackage{graphicx}

\usepackage{subfigure}
\usepackage{caption}
\usepackage{textcomp}

\usepackage{booktabs}
\usepackage{multirow}
\usepackage{wrapfig}
\usepackage{graphicx}
\usepackage{enumitem}
\usepackage{soul}
\usepackage{amsmath,amscd,amsbsy,amssymb,amsfonts,latexsym,url,bm,amsthm}
\def\BibTeX{{\rm B\kern-.05em{\sc i\kern-.025em b}\kern-.08em
		T\kern-.1667em\lower.7ex\hbox{E}\kern-.125emX}}
\usepackage{algorithm}
\usepackage{algorithmic}
\usepackage[algo2e,ruled,vlined,boxed,linesnumbered]{algorithm2e}
\usepackage{threeparttable}
\usepackage{listings}
\usepackage{colortbl}

\usepackage{multirow}
\usepackage{multicol}
\usepackage[colorlinks,
            linkcolor=black,
            anchorcolor=blue,
            citecolor=gray,
            ]{hyperref}
            
\usepackage{makecell}

\usepackage[capitalize,noabbrev]{cleveref}

\theoremstyle{plain}
\newtheorem{theorem}{Theorem}[section]

\theoremstyle{definition}

\theoremstyle{remark}

\newcommand{\model}{\textsc{Glind}\xspace}
\newcommand{\modelgcn}{\textsc{Glind-gcn}\xspace}
\newcommand{\modelgat}{\textsc{Glind-gat}\xspace}
\newcommand{\modeltrans}{\textsc{Glind-Trans}\xspace}

\newcommand{\std}[1]{{ #1}}

\usepackage[textsize=tiny]{todonotes}

\definecolor{first}{RGB}{178,24,43}
\definecolor{second}{RGB}{117,112,179}
\definecolor{third}{RGB}{189,189,189}
\icmltitlerunning{Learning Divergence Fields for Shift-Robust Graph Representations}

\begin{document}

\twocolumn[

\icmltitle{Learning Divergence Fields for Shift-Robust Graph Representations}



\icmlsetsymbol{equal}{*}

\begin{icmlauthorlist}
\icmlauthor{Qitian Wu}{sjtucs}
\icmlauthor{Fan Nie}{sjtucs}
\icmlauthor{Chenxiao Yang}{sjtucs}
\icmlauthor{Junchi Yan}{sjtucs}
\end{icmlauthorlist}

\icmlaffiliation{sjtucs}{School of Artificial Intelligence \&  Department of Computer Science and Engineering \& MoE Lab of AI, Shanghai Jiao Tong University, Shanghai, China}

\icmlcorrespondingauthor{Junchi Yan}{yanjunchi@sjtu.edu.cn}

\icmlkeywords{Machine Learning, ICML}

\vskip 0.3in
]



\printAffiliationsAndNotice{}  

\begin{abstract}
    Real-world data generation often involves certain geometries (e.g., graphs) that induce instance-level interdependence. This characteristic makes the generalization of learning models more difficult due to the intricate interdependent patterns that impact data-generative distributions and can vary from training to testing. In this work, we propose a geometric diffusion model with learnable divergence fields for the challenging generalization problem with interdependent data. We generalize the diffusion equation with stochastic diffusivity at each time step, which aims to capture the multi-faceted information flows among interdependent data. Furthermore, we derive a new learning objective through causal inference, which can guide the model to learn 
    generalizable patterns of interdependence that are insensitive across domains. Regarding practical implementation, we introduce three model instantiations that can be considered as the generalized versions of GCN, GAT, and Transformers, respectively, which possess advanced robustness against distribution shifts. We demonstrate their promising efficacy for out-of-distribution generalization on diverse real-world datasets. Source codes are available at \url{https://github.com/fannie1208/GLIND}.
\end{abstract}

\section{Introduction}\label{sec-intro}

Learning from data involving certain geometries is a fundamental challenge in machine learning~\cite{tenenbaum2000global,belkin2003laplacian,belkin2006manireg}. One common scenario entails explicit graph structures, where the observed edges create interdependence among data points; another more challenging scenario involves implicit structures, where interdependence also exists but is not directly observable from data. In both scenarios, due to the interdependence of data points, the commonly used i.i.d. assumption for modeling becomes invalid. Moreover, given the dynamic environment where the model interacts in the open-world setting, the training and testing data are often generated from different distributions, necessitating generalization under distribution shifts~\cite{ood-classic-1,ood-classic-2,ood-classic-3,invariant-trans,irm}. 

\begin{figure}
    \centering
    \includegraphics[width=0.8\linewidth]{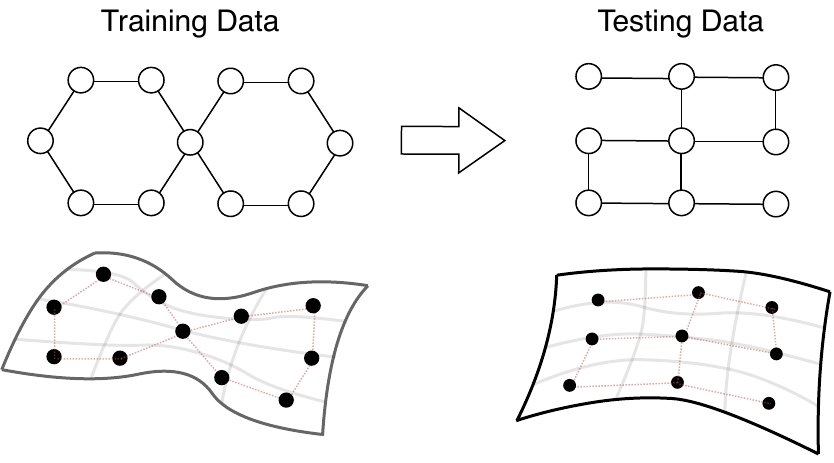}
    \vspace{-5pt}
    \caption{The challenge of generalization with interdependent data involves distribution shifts regarding the underlying manifolds that define the proximity among data samples.}
    \vspace{-10pt}
    \label{fig:problem}
\end{figure}

However, unlike standard learning settings (e.g., for image and text data) where each data sample can be treated as independent~\cite{ood-bench1,ood-bench2}, the interdependence among data points (e.g., nodes in an observed graph or other geometries with implicit structures) significantly increases the difficulty of generalization~\cite{ma2021subgroup,eerm,GKDE-2020,GPN-2021}. Particularly, as illustrated in Fig.~\ref{fig:problem},
since the training and testing data lie on different underlying manifolds that induce proximity among data points, the model solely optimized on training data with specific geometry may not generalize well to testing data generated from a different manifold. In contrast, an ideal model should be able to capture the \emph{generalizable} patterns of data interdependence that can transfer across domains (or interchangeably, environment). 

While generalization across potentially different environments is desirable in practice, the challenge lies in how to model the generalizable patterns of the interdependence among data points, which can be abstract and convoluted in nature. Given the data interdependence, the label of each instance depends on both the instance itself and other instances, and such input-label dependency, which is important for prediction, would vary accordingly when distribution shifts exist (between training and testing data). Thereby, for generalization, the model needs to learn causal predictive relations from the inputs of interdependent data with certain geometries to their labels, which stays insensitive to distribution shifts.


In this paper, we commence with diffusion equations on manifolds~\cite{weickert1998anisotropic,romeny2013geometry} as the foundation for learning generalizable predictive relations with interdependent data. We propose a geometric diffusion model with learnable divergence fields as an effective means to model the complex interdependent patterns among data points. In particular, we generalize the \emph{diffusivity} function, which measures the rate of information flows between two points on the manifold, as random samples from a variational distribution over a set of diffusivity hypotheses, inducing stochastic evolutionary directions at each step. In this way, the model can accommodate and capture the multi-faceted information flows at each diffusion step. 

To facilitate the generalization, we harness deconfounded learning, a technique from causal inference, and derive a step-wise re-weighting regularization approach for the inferred latent diffusivity. This gives rise to a new optimization objective that guides the model to learn the causal relation induced by the diffusion dynamics trajectory, which captures the generalizable relations from initial states ($x$) to output states ($y$) that are insensitive to data distribution shifts. In terms of practical implementation, we present three instantiations based on our model formulation, which can be treated as generalized versions of GCN~\citep{GCN-vallina}, GAT~\citep{GAT}, and DIFFormer~\citep{wu2023difformer}, respectively, for out-of-distribution generalization in situations where the data geometries are observed (e.g., as graphs) or unobserved.

We evaluate the practical efficacy of our models on diverse experimental datasets. The results show that the proposed models can effectively handle various distribution shifts.
The contributions are summarized below.

\textbf{$\circ$}\; We propose a geometric diffusion model with learnable divergence fields that accommodates multi-faceted information flows among data points. The diffusivity at each step is estimated via a variational distribution conditioned on the diffusion dynamics trajectories.

\textbf{$\circ$}\; We design a causal regularization approach for learning the generalizable patterns of interdependence. Our scheme serves as the first attempt to unlock the potential of diffusion models for generalization with interdependent data.

\textbf{$\circ$}\; We introduce three practical versions of model implementation and demonstrate their superiority through extensive experiments, particularly in handling various distribution shifts with observed and unobserved data geometries.

\section{Background}

In this section, we review some technical background as building blocks of our proposed model. 

\textbf{Diffusion on Manifolds.} The diffusion process over some abstract domain $\Omega$ composed of $N$ points describes the evolution of particular signal $z(u, t)$, a scalar-valued function on $\Omega \times [0, \infty)$, for each point $u\in \Omega$ at arbitrary time $t$. In specific, the evolution of $z(u, t)$ by a (heat) diffusion process can be described via a partial differential equation (PDE) with boundary conditions~\citep{freidlin1993diffusion,medvedev2014nonlinear,romeny2013geometry}:
\begin{equation}\label{eqn-heat-diff}
    \frac{\partial z(u, t)}{\partial t} = \nabla^*\left( D(u, t) \odot \nabla  z(u, t)\right),
\end{equation}
with initial conditions $z(u, 0) = z_0(u), t\geq 0, u\in \Omega$. Here $\nabla$, $\nabla^*$ and $D(u, t)$ reflect some spatial characteristics and are associated with the structure of $\Omega$. Without loss of generality, we assume $\Omega$ as a Riemannian manifold and define $\mathcal Z(\Omega)$ and $\mathcal Z(T\Omega)$ as the scalar and (tangent) vector fields on $\Omega$, respectively. Then the gradient operator $\nabla: \mathcal Z(\Omega) \rightarrow \mathcal Z(T\Omega)$ returns a vector field $\nabla z(u, t)$ that provides the direction of the steepest change of $z$ at the point $u$. The divergence operator $\nabla^*: \mathcal Z(T\Omega) \rightarrow \mathcal Z(\Omega)$ is a scalar field that reflects the flow of $\nabla z(u, t)$ through an infinitesimal volume around $u$. The $D(u, t)$ in Eqn.~\ref{eqn-heat-diff} is the \emph{diffusivity} that describes some thermal conductance properties of $\Omega$, i.e., the measure of the rate at which heat can spread over the space~\cite{rosenberg1997laplacian}. For homogeneous system, the diffusivity $D(u, t)$ remains a constant for arbitrary $u$, while $D(u, t)$ is position-dependent for inhomogeneous system. In the latter case, $D(u, t)$ can be scalar-valued (isotropic diffusion) or matrix-valued (anisotropic diffusion)~\cite{weickert1998anisotropic}.

\textbf{Causal Deconfounded Learning.} The principle of deconfounded learning is originally rooted in the study of Causality~\cite{causal-old1,causal-old2}, which pursuits the causal (a.k.a. stable) mappings between input $x$ (cause) and output $y$ (effect) and eliminates the influence from a latent confounder $c$, i.e., the common cause of $x$ and $y$. In machine learning tasks, the domain context in a dataset can be a common confounder between input features $x$ and labels $y$. For example, in image classification, the background often associates with both $x$ and $y$, e.g., a horse often appears on the grass. The domain context would mislead the model to learn non-stable correlation between inputs and labels, e.g., a shortcut between the `green' background and the `horse'~\cite{beery2018recognition,geirhos2018imagenet}. Such correlation however does not hold on other datasets collected with different contexts and may impair the model generalization. Deconfounded learning aims to remove the confounding bias from domain context by intervention on $x$ and maximize $p(y|do(x))$ where the $do$-operation cuts off the dependence from $c$ and $x$. The probability with the $do$ operator can be computed by randomized controlled trial using data recollection or statistically estimated by leveraging backdoor adjustment~\cite{causal-old1,christakopoulou2020deconfounding,roberts2020adjusting,yang2022temporal}. 



\section{Model Formulation}\label{sec-formulation}

We consider $N$ data samples (a.k.a. instances) $\{\mathbf x_u, y_u\}_{u=1}^N$ that are partially labeled and the labeled portion is $\{\mathbf x_u, y_u\}_{u=1}^M$ where $M<N$. There exist interconnecting structures among the $N$ data points, reflected by $\mathcal G=\{a_{uv}\}_{N\times N}$ where $a_{uv}=1$ indicates the connectivity between instance $u$ and $v$ and 0 for non-connectivity. In this situation, each instance $u$ can be considered as the node in the graph $\mathcal G$. Furthermore, without loss of generality, the structures can also be unobserved, in which case we can assume $a_{uv}=1$ for $\forall u, v$ suggesting that there can exist potential interactions between every instance pair. We introduce $\mathbf Z = [\mathbf z_u]_{u=1}^N$ where $\mathbf z_u\in \mathbb R^d$ denotes the representation (a.k.a. embedding) of instance $u$.

\subsection{Geometric Diffusion Model}

\textbf{Diffusion Equation over Discrete Space.} We first characterize a neural message passing model induced by generic diffusion process~Eqn.~\ref{eqn-heat-diff} defined over a discrete space consisting of $N$ points as locations on the manifold.
Through treating the instance representations as signals of locations in the discrete space, we can define the gradient and divergence operators according to the discretization of the continuous notions. The gradient operator $\nabla$ measures the difference between source and target locations, i.e., $(\nabla \mathbf Z(t))_{uv} = \mathbf z_u(t) - \mathbf z_v(t)$. The divergence operator $\nabla^*$ aggregates the information flows at a point, i.e., $(\nabla^*)_u = \sum_{v, a_{uv}=1} \mathbf d_{v}(\mathbf Z(t), u, t) \left(\nabla \mathbf Z(t)\right)_{uv}$, where $\mathbf d(\mathbf Z(t), u, t)$ determines the diffusivity of the location $u$ at time $t$. We can then obtain the diffusion equation that describes the evolution of instance representations over the observed discrete structures among $N$ data points:
\begin{equation}\label{eqn-diff-dis}
    \frac{\partial \mathbf z_u(t)}{\partial t} = \sum_{v, a_{uv}=1} \mathbf d_{v}(\mathbf Z(t), u, t) \left( \mathbf z_v(t) - \mathbf z_u(t)\right ),
\end{equation}
with initial conditions $\mathbf Z(0) = [\mathbf x_u]_{u=1}^N$ and $t\geq 0$. The above equation can be essentially considered as a continuous version of neural message passing, commonly used by graph neural networks~\cite{grand} and Transformers~\cite{wu2023difformer}, while the latter models uses discrete layers as the approximation of time. On top of this connection, recent works harness diffusion equations on graphs as a principled perspective for justifying architectural choices~\cite{beltrami,GRAND++}, analyzing GNNs' behaviors~\cite{wu2023advective} and guiding model designs for difficult problems, e.g., geometric knowledge distillation~\cite{geokd}.

\begin{figure*}
    \centering
    \includegraphics[width=\textwidth]{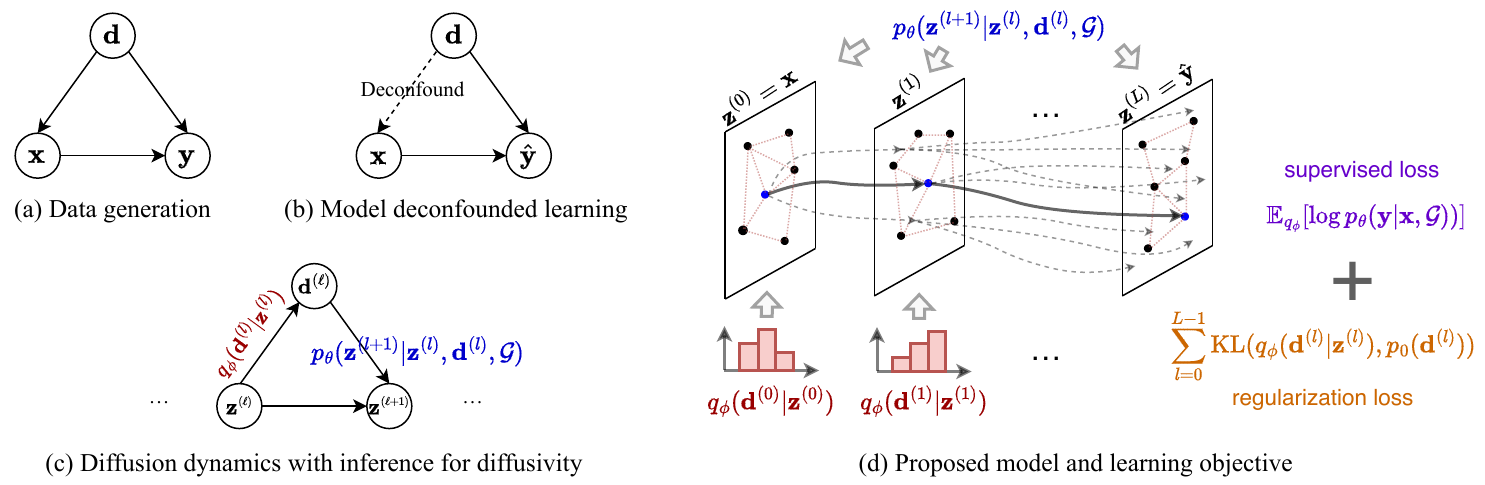}
    \caption{(a)$\sim$ (c) Dependence among random variables of interest (here we omit $\mathcal G$ for brevity since everything can be treated as conditioned on $\mathcal G$). (a) Causal dependence for data generation where the diffusivity $\mathbf d$ is the common cause of $\mathbf x$ and $\mathbf y$. (b) Deconfounded learning $p_\theta(\mathbf y|do(\mathbf x), \mathcal G)$ which aims to cut off the dependence path from $\mathbf d$ to $\mathbf x$ in order to learn causal (a.k.a. stable) relation between $\mathbf x$ and $\mathbf y$ for generalization. (c) Our diffusion model whose feed-forward dynamics $\mathbf x_u = \mathbf z_u^{(0)} \rightarrow \mathbf z_u^{(1)} \rightarrow \cdots \rightarrow \mathbf z_u^{(L)} = \hat {\mathbf y}_u$ is given by a predictive distribution $p_\theta(\mathbf z^{(l+1)} | \mathbf z^{(l)}, \mathbf d^{(l)}, \mathcal G)$ and a variational distribution $q_\phi(\mathbf d^{(l)}| \mathbf z^{(l)})$. (d) The geometric diffusion model that is optimized by a new learning objective (comprised of a supervised term and a regularization term) which can achieve the goal of deconfounded learning $p_\theta(\mathbf y|do(\mathbf x), \mathcal G)$.}
    \label{fig:model}
\end{figure*}

\textbf{Branching-Structured Divergence Fields.} The above diffusion system assumes that at each time step the information flows between connected locations are determined by a deterministic diffusivity function $\mathbf d(\mathbf Z(t), u, t)$. The diffusivity essentially measures the rate of information flows between any pair of locations on the manifold. Based on the analogy between diffusion on manifolds and message passing among interconnected data, the diffusivity can be treated as quantification of pairwise influence among data points. The latter, however, could be driven by multiple criteria along with uncertainty, instead of remaining to be deterministic. In light of this observation, we extend Eqn.~\ref{eqn-diff-dis} to accommodate the heterogeneity and stochasticity of the diffusivity, which is assumed to be sampled from a generative distribution $p(\mathbf d^{(t)} |\mathbf Z(t), u, t)$, over a set of hypothesis, dependent on the current states $\mathbf Z(t)$, position $u$ and time $t$:
\begin{equation}\label{eqn-diff-ours}
    \frac{\partial \mathbf z_u(t)}{\partial t} = \sum_{v, a_{uv}=1}   d_{uv}^{(t)} \cdot \left( \mathbf z_v(t) - \mathbf z_u(t)\right ),
\end{equation}
where $[d_{uv}^{(t)}]_{v=1}^N = \mathbf d_u^{(t)} \sim p(\mathbf d^{(t)}|\mathbf Z(t), u, t)$. Through the diffusion process defined by Eqn.~\ref{eqn-diff-ours}, the change of instance representation at infinitesimal time is equal to the \emph{divergence} at each point $u$ induced by the current diffusivity $\mathbf d_u^{(t)}$. The evolutionary direction of instance representation at time $t$ is stochastic according to $p(\mathbf d^{(t)}|\mathbf Z(t), u, t)$ that yields multiple branches of subsequent diffusion dynamics from the current time (as illustrated in Fig.~\ref{fig:model}(d)). This formulation further paves the way for generalization by enabling the model to learn the stable diffusion dynamics as we will discuss in later subsections.

\textbf{Diffusion Trajectories with Stochastic Diffusivity.} The differential equation of Eqn.~\ref{eqn-diff-ours} can be solved through numerical methods by using discrete time steps $\{l\}$ to approximate the continuous time $t$. In particular, with step size $\alpha$, Eqn.~\ref{eqn-diff-ours} induces a trajectory of instance embeddings:
\begin{equation}\label{eqn-diff-traj}
    \mathbf z_u^{(l+1)} = \mathbf z_u^{(l)} + \alpha \sum_{v, a_{uv}=1}   d_{uv}^{(l)} \cdot \left( \mathbf z_v^{(l)} - \mathbf z_u^{(l)} \right ),
\end{equation}
where $[d_{uv}^{(t)}]_{v=1}^N = \mathbf d_u^{(l)} \sim p(\mathbf d^{(l)}|\mathbf z^{(l)}_u)$. The diffusion model Eqn.~\ref{eqn-diff-traj} with $L$ layers yields a trajectory of embeddings for each instance $u$: $\mathbf x_u = \mathbf z_u^{(0)} \rightarrow \mathbf z_u^{(1)} \rightarrow \cdots \rightarrow \mathbf z_u^{(L)} = \hat {\mathbf y}_u$, where $\hat {\mathbf y}_u$ denotes the predicted label for $u$.

\subsection{Optimization with Causal Regularization}

For generalization, we next discuss on how to guide the geometric diffusion model to learn causal relations from inputs to outputs that are insensitive to distribution shifts. Consider the instance embedding $\mathbf z_u^{(l)}$ of an arbitrary given layer as a random variable, we define by $p_\theta(\mathbf z^{(l+1)} | \mathbf z^{(l)}, \mathbf d^{(l)}, \mathcal G)$ the (delta) predictive distribution induced by the $l$-th layer's updating of Eqn.~\ref{eqn-diff-traj}, where the condition on $\mathcal G$ stems from the message passing among instances. Then the generative distribution at the $l$-th layer can be expressed as 
\begin{equation}\nonumber
    p_\theta(\mathbf z^{(l+1)} | \mathbf z^{(l)}, \mathcal G) = \mathbb E_{p(\mathbf d^{(l)} | \mathbf z^{(l)})} [ p_\theta(\mathbf z^{(l+1)} | \mathbf z^{(l)}, \mathbf d^{(l)}, \mathcal G) ] ,
\end{equation}
where the expectation marginalizes over all the possible results of $\mathbf d^{(l)}$. The log-likelihood $\log p_\theta(\mathbf y|\mathbf x, \mathcal G)$ for each data sample can be computed by $\log \prod_{l=0}^{L-1} p_\theta(\mathbf z^{(l+1)} | \mathbf z^{(l)}, \mathcal G) =$
\begin{equation}\label{eqn-obj-ori}
    \sum_{l=0}^{L-1} \log \mathbb E_{p(\mathbf d^{(l)} | \mathbf z^{(l)})} [ p_\theta(\mathbf z^{(l+1)} | \mathbf z^{(l)}, \mathbf d^{(l)}, \mathcal G) ].
\end{equation}

\textbf{Inference for Diffusivity.} Due to the intractable integration over $\mathbf d^{(l)}$ in the above objective, we turn to the evidence lower bound of Eqn.~\ref{eqn-obj-ori} by introducing a variational distribution $q_\phi(\mathbf d^{(l)} | \mathbf z^{(l)})$ (where $\phi$ denotes its parameterization):
\begin{equation}\label{eqn-obj-var}
    \sum_{l=0}^{L-1} \mathbb E_{q_\phi(\mathbf d^{(l)} | \mathbf z^{(l)} )} \left [ \log p_\theta(\mathbf z^{(l+1)} | \mathbf z^{(l)}, \mathbf d^{(l)}, \mathcal G) \frac{p(\mathbf d^{(l)} | \mathbf z^{(l)} )}{q_\phi(\mathbf d^{(l)} | \mathbf z^{(l)} )} \right ].
\end{equation}
With some re-arranging, the evidence lower bound of the log-likelihood $\log p_\theta(\mathbf y|\mathbf x, \mathcal G)$ induces the following objective  (see derivation in Appendix~\ref{appx-prove-1})
\begin{align}\label{eqn-obj-var2}
    & \mathbb E_{q_\phi(\mathbf d^{(0)} | \mathbf z^{(0)}), \cdots, q_\phi(\mathbf d^{(L-1)} | \mathbf z^{(L-1)} )} \left [\log p_\theta (\mathbf y|\mathbf x, \mathbf d^{(0)}, \cdots, \right. \nonumber \\
    & \left. \mathbf d^{(L-1)}, \mathcal G) \right ] 
    - \sum_{l=0}^{L-1} \mbox{KL}(q_\phi(\mathbf d^{(l)} | \mathbf z^{(l)} ), p(\mathbf d^{(l)} | \mathbf z^{(l)} )),
\end{align}
where the first term corresponds to the supervised loss by computing the difference between the ground-truth label $\mathbf y_u$ and the prediction $\hat{\mathbf y}_u$, and the second term has a regularization effect on the inferred latent diffusivity. Eqn.~\ref{eqn-obj-var2} is a tractable objective that approximates the log-likelihood. 

However, the generative distribution $p(\mathbf d^{(l)} | \mathbf z^{(l)})$ is a model-based prior for diffusivity over a hypothesis space that expresses how plausible the model thought the ideal diffusivity were in light of the previous diffusion dynamics. Though it incorporates the information of $\mathbf z^{(l)}$ for estimating the adaptive diffusivity at the current step, the model learning would make the estimation of diffusivity biased towards training data.  
In this way, the predictive relations from $\mathbf z^{(l)}$ (resp. $\mathbf x$) to $\mathbf z^{(l+1)}$ (resp. $\mathbf y$) learned by the model would be spuriously associated by the latent confounder, the diffusivity $\mathbf d^{(l)}$ (resp. $\{\mathbf d^{(l)}\}_{l=0}^{L-1}$), as illustrated by Fig.~\ref{fig:model}(a). The confounding effect of the diffusivity, which corresponds to the context information that differs across domains, would hamper the generalization of the model when transferring from the training distribution to a testing one where the data lies on another manifold with disparate proximity context.

\textbf{Step-wise Re-weighting Regularization} To facilitate generalization, the ideal solution is to achieve deconfounded learning, i.e., $p_\theta(\mathbf y|do(\mathbf x), \mathcal G)$ that cuts off the dependence path from $\mathbf d$ to $\mathbf x$ (as illustrated in Fig.~\ref{fig:model}(b)), in which situation the model can learn the causal (a.k.a. stable) relation between $\mathbf x$ and $\mathbf y$ that is generalizable across domains. While the technical difficulty lies in how to compute $p_\theta(\mathbf y|do(\mathbf x), \mathcal G)$ in a feasible manner, the next theorem presents a tractable surrogate that approximates the target. 
\begin{theorem}\label{thm-1}
For any given diffusion model $p_\theta(\mathbf z^{(l+1)}|\mathbf z^{(l)}, \mathbf d^{(l)}, \mathcal G)$, we have a lower bound of the deconfounded learning objective: $\log p_\theta(\mathbf y|do(\mathbf x), \mathcal G) \geq $
\vspace{-5pt}
\begin{equation}\label{eqn-obj-reweight}
    \sum_{l=0}^{L-1} \mathbb E_{q_\phi(\mathbf d^{(l)} | \mathbf z^{(l)} )} \left [ \log p_\theta(\mathbf z^{(l+1)} | \mathbf z^{(l)}, \mathbf d^{(l)}, \mathcal G) \frac{p_0(\mathbf d^{(l)} )}{q_\phi(\mathbf d^{(l)} | \mathbf z^{(l)} )} \right ],
    \end{equation}
where $p_0(\mathbf d^{(l)})$ is a model-free prior distribution. In particular, the equality holds for Eqn.~\ref{eqn-obj-reweight} iff $q_\phi(\mathbf d^{(l)} | \mathbf z^{(l)} ) = p_\theta(\mathbf d^{(l)} | \mathbf z^{(l)}, \mathbf z^{(l+1)}, \mathcal G) \cdot \frac{p_0(\mathbf d^{(l)})}{p(\mathbf d^{(l)} | \mathbf z^{(l)})}$.
\end{theorem}
The proof is deferred to Appendix~\ref{appx-prove-2}. We can optimize Eqn.~\ref{eqn-obj-reweight} as a surrogate objective by joint learning of $q_\phi$ and $p_\theta$ that has the effect of optimizing the deconfounded learning objective.
As another perspective for understanding the effect of Eqn.~\ref{eqn-obj-reweight}, it can be seen as multiplying a re-weighting term $\frac{p_0(\mathbf d^{(l)} )}{p(\mathbf d^{(l)} | \mathbf z^{(l)} )}$ with Eqn.~\ref{eqn-obj-var}, which can down-weight the frequent diffusivity components and up-weight the infrequent ones to eliminate the observation bias caused by limited training data. This can facilitate learning the stable relation from $\mathbf z^{(l)}$ (resp. $\mathbf x$) to $\mathbf z^{(l+1)}$ (resp. $\mathbf y$) that holds across domains, instead of the non-stable correlation pertaining to specific context in training data. 

Based on Eqn.~\ref{eqn-obj-reweight}, we can obtain the final learning objective (see derivation in Appendix~\ref{appx-prove-3} which is similar to Eqn.~\ref{eqn-obj-var2}):
\vspace{-10pt}
\begin{align}\label{eqn-obj-reweight2}
    & \mathbb E_{q_\phi(\mathbf d^{(0)} | \mathbf z^{(0)} ), \cdots, q_\phi(\mathbf d^{(L-1)} | \mathbf z^{(L-1)} )} \left [\log p_\theta (\mathbf y|\mathbf x, \mathbf d^{(0)}, \cdots, \right. \nonumber \\
    &\left. \mathbf d^{(L-1)}, \mathcal G) \right ] - \sum_{l=0}^{L-1}\mbox{KL}(q_\phi(\mathbf d^{(l)} | \mathbf z^{(l)} ), p_0(\mathbf d^{(l)})).
\end{align}
Fig.~\ref{fig:model}(d) provides an illustration for the model. The computation of the regularization term depends on specific form of $q_\phi(\mathbf d^{(l)} | \mathbf z^{(l)} )$, which we will discuss in the next section.

\section{Model Instantiations}\label{sec-model}

In this section, we delve into the model instantiations on the basis of the formulation of our diffusion model in Sec.~\ref{sec-formulation}. We call our model \textbf{\model}, short for diffusion-induced \textbf{G}eneralizable \textbf{L}earning with \textbf{IN}terdependent \textbf{D}ata. For simplicity, we assume the inference distribution for diffusivity $q_\phi(\mathbf d^{(l)} |\mathbf z^{(l)})$ as a multinomial distribution: at each time step $l$ for arbitrary instance $u$, the diffusivity $\mathbf d_u^{(l)}$ is sampled from a limited hypothesis set $\{\mathbf d^{(l,k)}_u\}_{k=1}^K$ with probabilities $\{\pi_{u}^{(l, k)}\}_{k=1}^K$ for each component, where $\sum_{k=1}^K \pi_u^{(l,k)}=1$. 
We can further assume $\alpha=1$, and then the diffusion trajectory of Eqn.~\ref{eqn-diff-traj} induces a feed-forward propagation layer:
\begin{equation}\nonumber
    \mathbf z_u^{(l+1)} = \mathbf z_u^{(l)} + \sum_{k=1}^K h_{u,k}^{(l)} \sum_{v, a_{uv}=1} d^{(l,k)}_{uv} (\mathbf z_v^{(l)} - \mathbf z_u^{(l)}),
\end{equation}
where $\mathbf h_u^{(l)}\sim \mathcal M(\bm \pi_{u}^{(l)})$ is a one-hot vector where the entry `1' indicates the sampled result from the multinomial distribution $\mathcal M(\bm \pi_{u}^{(l)})$. Here $\bm \pi_u^{(l)}$ and $\mathbf d_{u}^{(l,k)}$ determine the divergence (the updating signals) at the current time step $l$ and control the evolutionary direction of the diffusion process, and both of them can be parameterized.

\textbf{Parameterization.} We adopt a feed-forward layer with Softmax to model the probability of $K$ diffusivity components and leverage the Gumbel trick~\cite{gumbel-iclr17} to approximate the sampling process:
\begin{equation}\nonumber
    h_{u,k}^{(l)} = \frac{\exp\left(\left(\pi^{(l,k)}_{u}+g_k\right)/\tau\right)}{\sum_{k'}\exp((\pi^{(l,k')}_{u}+g_{k'})/\tau)}, ~~ g_k \sim \mbox{Gumbel}(0,1),
\end{equation}
where $[\pi^{(l,k)}_{u}]_{k=1}^K = \boldsymbol \pi^{(l)}_u = \mbox{Softmax}(\mathbf W_L^{(l)} \mathbf z_u^{(l)})$ and $\mathbf W_L^{(l)}\in \mathbb R^{K\times d}$ is a trainable weight matrix. The message passing is driven by the diffusivity hypothesis set $\{\mathbf d^{(l,k)}_u\}_{k=1}^K$. We next introduce three model instantiations that can be considered as the generalized implementations of GCN, GAT and Transformers, respectively.

\textbf{\modelgcn.} We first consider a model version that uses the aggregated embeddings of connected instances to update the representation of the target instance. Inspired by the anisotropic diffusion which is position- and direction-dependent, we assume each branch $k$ converts the embeddings for the target instance (resp. other connected instances) through $\mathbf W_S^{(l,k)}\in \mathbb R^{d\times d}$ (resp. $\mathbf W_D^{(l,k)}\in \mathbb R^{d\times d}$) and the gradient of signals is scaled by the number of connected instances $\tilde d_u = \sum_{v} a_{uv}$:

\vspace{-15pt}
\scriptsize{
\begin{equation}\nonumber
    \mathbf z_u^{(l+1)} = \mathbf z_u^{(l)} + \sum_{k=1}^K h_{u,k}^{(l)} \left (\sum_{v,a_{uv}=1}  \frac{1}{\tilde d_u} \mathbf W_D^{(l,k)} \mathbf z_v^{(l)} + \mathbf W_S^{(l,k)} \mathbf z_u^{(l)} \right ),
\end{equation}
}
\normalsize
where $\{\mathbf W_S^{(l,k)}\}_{k=1}^K$ and $\{\mathbf W_D^{(l,k)}\}_{k=1}^K$ are trainable weights at the $l$-th layer. This model can be seen as a generalized implementation of the GCN architecture~\cite{GCN-vallina}, where the approximate one-hot vector $\mathbf h_{u}^{(l)}$ dynamically selects the convolution filter in each layer. 

\textbf{\modelgat.} We can further harness an attention network for each branch to model the intensity of the pairwise influence between connected nodes, by extending the spirit of GAT~\cite{GAT}:

\scriptsize{
\vspace{-15pt}
\begin{equation}\nonumber
    \mathbf z_u^{(l+1)} = \mathbf z_u^{(l)} + \sum_{k=1}^K h_{u,k}^{(l)} \left ( \sum_{v, a_{uv}=1} w^{(l,k)}_{uv}  \mathbf W_D^{(l,k)} \mathbf z_v^{(l)} + \mathbf W_S^{(l,k)} \mathbf z_u^{(l)} \right ),
\end{equation}
\begin{equation}\nonumber
    w^{(l,k)}_{uv} = \frac{\delta((\mathbf c^{(l,k)})^\top[\mathbf W_A^{(l,k)} \mathbf z_u^{(l)} \| \mathbf W_A^{(l,k)} \mathbf z_v^{(l)} ])}{\sum_{w, a_{uw}=1} \delta(\mathbf c^{(l,k)})^\top[\mathbf W_A^{(l,k)} \mathbf z_u^{(l)} \| \mathbf W_A^{(l,k)} \mathbf z_w^{(l)} ])},
\end{equation}
}
\normalsize
where $\delta$ is instantiated as LeakyReLU and $\mathbf W_A^{(l,k)}\in \mathbb R^{d\times d}$ and $\mathbf c^{(l,k)} \in \mathbb R^{2d}$ are both trainable parameters. Here, $\mathbf h_{u}^{(l)}$ adaptively selects the path for attentive propagation. 

\textbf{\modeltrans.} The preceding two models resort to message passing over observed structures. There also exist scenarios where the data geometries are partially observed or unobserved. To accommodate the potential interactions among arbitrary instance pairs, we extend the attention-based propagation to a latent topology as is done by Transformer~\cite{vaswani2017attention}: 
\begin{equation}\nonumber
    \mathbf z_u^{(l+1)} = \mathbf z_u^{(l)} + \sum_{k=1}^K h_{u,k}^{(l)} \left (\mathbf W_D^{(l,k)} \mathbf b_{u}^{(l, k)} + \mathbf W_S^{(l,k)} \mathbf z_u^{(l)} \right ),
\end{equation}
\begin{equation}\nonumber
    \mathbf b_{u}^{(l, k)} = \sum_{v} \frac{\eta(\mathbf W_K^{(l, k)}\mathbf z_v^{(l)}, \mathbf W_Q^{(l, k)}\mathbf k_u^{(l)}) }{\sum_{w=1}^N \eta(\mathbf W_K^{(l, k)}\mathbf z_w^{(l)}, \mathbf W_Q^{(l, k)}\mathbf k_u^{(l)})} \cdot \mathbf z_v^{(l)},
\end{equation}
where $\mathbf W_K^{(l, k)}$ and $\mathbf W_Q^{(l, k)}$ are trainable weights at the $l$-th layer, and $\eta$ denotes a certain similarity function. However, due to the all-pair global attention, the above computation requires quadratic complexity w.r.t. $N$, which hinders the scalability to large numbers of instances. We thus turn to a linearly scalable and numerically stable version of all-pair attention proposed by DIFFormer~\cite{wu2023difformer}, which specifically instantiates $\eta$ as $\eta(\mathbf a, \mathbf b) = 1 + (\frac{\mathbf a}{\|\mathbf a\|_2} )^\top \frac{\mathbf b}{\|\mathbf b\|_2}$. Then assuming $\mathbf k^{(l)}_u = \frac{\mathbf W_K^{(l, k)}\mathbf z_u^{(l)}}{\|\mathbf W_K^{(l, k)}\mathbf z_u^{(l)} \|_2}$ and $\mathbf q^{(l)}_u = \frac{\mathbf W_Q^{(l, k)}\mathbf z_u^{(l)}}{\|\mathbf W_Q^{(l, k)}\mathbf z_u^{(l)} \|_2}$, we can efficiently compute $\mathbf b_{u,v}^{(l, k)}$ by 
\begin{equation}\nonumber
    \mathbf b_{u}^{(l, k)} = \frac{\sum_{v=1}^N \mathbf z_v^{(l)} + \left ( \sum_{v=1}^N (\mathbf k_v^{(l)}) (\mathbf z_v^{(l)})^\top  \right ) (\mathbf q_u^{(l)} ) }{N + (\mathbf q_u^{(l)})^\top (\sum_{v=1}^N \mathbf k_v^{(l)}) }.
\end{equation}
The above computation for updating $N$ instances' embeddings in each layer can be achieved within $O(N)$ complexity since the three summation terms are shared by all instances and only requires once computation in practice. Moreover, if the observed structures are available, we can incorporate the aggregated embeddings of connected instances to the updating signals: $\mathbf b_{u}^{(l, k)} \leftarrow \frac{1}{2}(\mathbf b_{u}^{(l, k)} + \sum_{v, a_{uv}=1} \frac{1}{\tilde d_u} \mathbf z_{v}^{(l)})$.

\paragraph{Data-Driven Prior via Mixture of Posteriors.}
A proper setting of the prior distribution $p_0(\mathbf d^{(l)})$ is important but non-trivial. One simple solution is to assume the prior as some pre-defined trivial forms~\cite{dinh2015nice}, which however may potentially lead to over-regularization~\cite{burda2015importance}. Alternatively, one can estimate the prior by the average of the variational posterior~\cite{hoffman2016elbo}, i.e., $p_0(\mathbf d^{(l)}) \approx 1/N \sum_{u=1}^N q(\mathbf d^{(l)}|\mathbf z^{(l)}=\mathbf z^{(l)}_u)$.
Nevertheless, such an approach can be computationally expensive~\cite{tomczak2018vae} and would lead to biased estimation given limited training data. In our case, inspired by~\cite{tomczak2018vae} using mixture of Gaussian as a learnable prior, we propose to use a mixture of pseudo variational posteriors as a flexible prior estimation:
\begin{equation}\label{eqn-prior}
    p_0(\mathbf d^{(l)}) = \frac{1}{T} \sum_{t=1}^T q(\mathbf d^{(l)}|\mathbf z^{(l)}=\tilde{\mathbf z}^{(l)}_t).
\end{equation}
where $\tilde{\mathbf z}^{(l)}_t$ is the embedding of instance $t$ in the generated pseudo dataset $\{\tilde{\mathbf x}_t, \tilde y_t \}_{t=1}^T$. The latter is constructed by randomly sampling $T$ instances from the $N$ instances in the observed dataset, and the structures of the pseudo dataset are randomly generated with a pre-defined probability for each potential edge. We set $T\ll N$ to reduce the computational cost. In this way, the prior estimation by Eqn.~\eqref{eqn-prior} plays as a general reflection of how the model recognizes each diffusivity hypothesis given uninformative inputs and is learned with the model in a fully data-driven manner. 

\begin{table*}[t!]
	\centering
	\caption{Testing (mean$\pm$standard deviation) Accuracy (\%) for \texttt{Arxiv} and ROC-AUC (\%) for \texttt{Twitch} on different subsets of out-of-distribution data (determined by publication times and subgraphs, respectively). The missing results of EERM on \texttt{Arxiv} stem from the out-of-memory issue. We mark the \textbf{\textcolor{first}{first}/\textcolor{second}{second}/\textcolor{third}{third}} methods with the top performance.}
	\label{tbl-res-observed}
	\begin{threeparttable}
        \setlength{\tabcolsep}{3mm}{ 
        \resizebox{\textwidth}{!}{
		\begin{tabular}{l|ccc|ccc}
		    \hline
 \multirow{2}{*}{\textbf{Method}} & \multicolumn{3}{c|}{\texttt{Arxiv}} & \multicolumn{3}{c}{\texttt{Twitch}} \\ 
        & \textbf{2014-2016} & \textbf{2016-2018} & \textbf{2018-2020} & \textbf{ES} & \textbf{FR} & \textbf{EN} \\  
 			\hline
    ERM-GCN & 56.33 $\pm$ 0.17 & 53.53\std{$\pm$ 0.44} & 45.83\std{$\pm$ 0.47} & 66.07\std{$\pm$ 0.14} & 52.62\std{$\pm$ 0.01} & \textcolor{third}{\textbf{63.15 $\pm$ 0.08}} \\
   IRM-GCN     & 55.92\std{$\pm$ 0.24} & 53.25\std{$\pm$ 0.49} & 45.66\std{$\pm$ 0.83} & \textcolor{third}{\textbf{66.95 $\pm$ 0.27}} & 52.53\std{$\pm$ 0.02} & 62.91\std{$\pm$ 0.08} \\
   GroupDRO-GCN & 56.52\std{$\pm$ 0.27} & 53.40\std{$\pm$ 0.29} & 45.76\std{$\pm$ 0.59} & 66.82\std{$\pm$ 0.26} & \textcolor{third}{\textbf{52.69 $\pm$ 0.02}} & 62.95\std{$\pm$ 0.11} \\
   DANN-GCN   & \textcolor{third}{\textbf{56.35\std{$\pm$ 0.11}}} & \textcolor{third}{\textbf{53.81\std{$\pm$ 0.33}}} & \textcolor{third}{\textbf{45.89\std{$\pm$ 0.37}}} & 66.15\std{$\pm$ 0.13} & 52.66\std{$\pm$ 0.02} & \textcolor{second}{\textbf{63.20 $\pm$ 0.06}} \\
    Mixup-GCN   & \textcolor{second}{\textbf{56.67\std{$\pm$ 0.46}}} & \textcolor{second}{\textbf{54.02\std{$\pm$ 0.51}}} & \textcolor{second}{\textbf{46.09\std{$\pm$ 0.58}}} & 65.76\std{$\pm$ 0.30} & \textcolor{second}{\textbf{52.78 $\pm$ 0.04}} & \textcolor{third}{\textbf{63.15 $\pm$ 0.08}} \\
  EERM-GCN & - & - & - & \textcolor{second}{\textbf{67.50 $\pm$ 0.74}}& 51.88\std{$\pm$ 0.07}& 62.56\std{$\pm$ 0.02}\\
  \textbf{\modelgcn} & \textcolor{first}{\textbf{59.42 $\pm$ 0.33}} & \textcolor{first}{\textbf{56.84 $\pm$ 0.54 }} & \textcolor{first}{\textbf{57.06 $\pm$ 1.21}} & \textcolor{first}{\textbf{67.72 $\pm$ 0.10}} & \textcolor{first}{\textbf{53.16 $\pm$ 0.08}} & \textcolor{first}{\textbf{64.18 $\pm$ 0.03}}\\
            \hline
ERM-GAT & 57.15\std{$\pm$ 0.25} & 55.07\std{$\pm$ 0.58} & 46.22\std{$\pm$ 0.82} & 65.67\std{$\pm$ 0.02} & 52.00\std{$\pm$ 0.10} & 61.85\std{$\pm$ 0.05}\\
 IRM-GAT & 56.55\std{$\pm$ 0.18} & 54.53\std{$\pm$ 0.32} & 46.01\std{$\pm$ 0.33} & \textcolor{third}{\textbf{67.27 $\pm$ 0.19}} & 52.85\std{$\pm$ 0.15} & \textcolor{third}{\textbf{62.40 $\pm$ 0.24}}\\
  GroupDRO-GAT & 56.69\std{$\pm$ 0.27} & 54.51\std{$\pm$ 0.49} & 46.00\std{$\pm$ 0.59} & \textcolor{second}{\textbf{67.41 $\pm$ 0.04}} & \textcolor{second}{\textbf{52.99 $\pm$ 0.08}} & 62.29\std{$\pm$ 0.03}\\
  DANN-GAT & \textcolor{second}{\textbf{57.23 $\pm$ 0.18}} & \textcolor{third}{\textbf{55.13 $\pm$ 0.46}} & \textcolor{third}{\textbf{46.61\std{$\pm$ 0.57}}} & 66.59\std{$\pm$ 0.38} & \textcolor{third}{\textbf{52.88 $\pm$ 0.12}} & \textcolor{second}{\textbf{62.47 $\pm$ 0.32}}\\
 Mixup-GAT & \textcolor{third}{\textbf{57.17 $\pm$ 0.33}} & \textcolor{second}{\textbf{55.33 $\pm$ 0.37}} & \textcolor{second}{\textbf{47.17 $\pm$ 0.84}} & 65.58\std{$\pm$ 0.13} & 52.04\std{$\pm$ 0.04} & 61.75\std{$\pm$ 0.13} \\
 EERM-GAT & - & - & - & 66.80\std{$\pm$ 0.46}& 52.39\std{$\pm$ 0.20}& 62.07\std{$\pm$ 0.68}\\
\textbf{\modelgat} & \textcolor{first}{\textbf{60.36 $\pm$ 0.36}} & \textcolor{first}{\textbf{58.98 $\pm$ 0.43}} & \textcolor{first}{\textbf{59.71 $\pm$ 0.53}} & \textcolor{first}{\textbf{67.82 $\pm$ 0.10}} & \textcolor{first}{\textbf{54.50 $\pm$ 0.12}} & \textcolor{first}{\textbf{64.32 $\pm$ 0.12}}\\
  \hline
		\end{tabular}}}
	\end{threeparttable}
\end{table*}

\begin{table*}[t!]
\centering
\caption{Testing RMSE for \texttt{DPPIN} on different domains (determined by protein identification methods).\label{tbl-res-dppin}}
\begin{threeparttable}
\centering
\setlength{\tabcolsep}{3mm}{ 
    \resizebox{\textwidth}{!}{
\begin{tabular}{l|ccccccc}
\hline

 \textbf{Method}  & \textbf{Hazbun} & \textbf{Krogan (LCMS)} & \textbf{Krogan (MALDI)} & \textbf{Lambert} & \textbf{Tarassov} & \textbf{Uetz} & \textbf{Yu}\\

\hline
ERM-Trans 	& 1.82 $\pm$ 0.17 & \textcolor{second}{\textbf{1.63 $\pm$ 0.04}} & \textcolor{second}{\textbf{1.57 $\pm$ 0.03}} & 1.49 $\pm$ 0.07 & \textcolor{second}{\textbf{1.62 $\pm$ 0.03}} & \textcolor{third}{\textbf{1.52 $\pm$ 0.04}} & \textcolor{third}{\textbf{1.51 $\pm$ 0.04}} \\
IRM-Trans  &  1.66 $\pm$ 0.14 & 1.86 $\pm$ 0.04 & 1.84 $\pm$ 0.04 & 1.52 $\pm$ 0.07 & 1.76 $\pm$ 0.03 &  1.66 $\pm$ 0.05 & 1.66 $\pm$ 0.04\\
DANN-Trans   & 1.69 $\pm$ 0.11 & \textcolor{third}{\textbf{1.66 $\pm$ 0.02}} & \textcolor{third}{\textbf{1.62 $\pm$ 0.03}} & \textcolor{second}{\textbf{1.39 $\pm$ 0.05}} & \textcolor{third}{\textbf{1.63 $\pm$ 0.01}} & \textcolor{second}{\textbf{1.49 $\pm$ 0.01}} & \textcolor{second}{\textbf{1.50 $\pm$ 0.01}}\\
GroupDRO-Trans   & \textcolor{third}{\textbf{1.65 $\pm$ 0.13}} & 1.68 $\pm$ 0.02 & 1.65 $\pm$ 0.02 & 1.48 $\pm$ 0.03 & 1.72 $\pm$ 0.01 & 1.53 $\pm$ 0.04 & 1.53 $\pm$ 0.01\\
Mixup-Trans    & \textcolor{second}{\textbf{1.46 $\pm$ 0.13}} & 1.79 $\pm$ 0.05& 1.76 $\pm$ 0.04& 1.50 $\pm$ 0.06& 1.70 $\pm$ 0.05& 1.56 $\pm$ 0.06 & 1.59 $\pm$ 0.06\\
EERM-Trans    & 1.68 $\pm$ 0.47 & 1.91 $\pm$ 0.23 & 1.92 $\pm$ 0.09 & \textcolor{third}{\textbf{1.47 $\pm$ 0.05}} & 1.79 $\pm$ 0.11 & 1.67 $\pm$ 0.07 & 1.65 $\pm$ 0.08\\
\textbf{\modeltrans} &  \textcolor{first}{\textbf{1.02 $\pm$ 0.07}} & \textcolor{first}{\textbf{1.38 $\pm$ 0.07}} & \textcolor{first}{\textbf{1.33 $\pm$ 0.05}} & \textcolor{first}{\textbf{1.08 $\pm$ 0.04}} & \textcolor{first}{\textbf{1.40 $\pm$ 0.04}} & \textcolor{first}{\textbf{1.20 $\pm$ 0.04}} & \textcolor{first}{\textbf{1.20 $\pm$ 0.04}}\\
\hline
\end{tabular} }}
\end{threeparttable}
\end{table*}

\section{Experiments}\label{sec-exp}

The goal of our experiments is to evaluate the generalization ability of the model under distribution shifts with interdependent data. We consider various real-world datasets that involve observed and unobserved data geometries, and the distribution shifts between training and testing sets are led by distinct contexts. Following common practice, we use the prediction accuracy on the out-of-distribution testing data for measuring the generalization performance.

\subsection{Experiment Setup}\label{sec-exp-setup}

\textbf{Datasets.} We adopt five datasets \texttt{Twitch}, \texttt{Arxiv}, \texttt{DPPIN}, \texttt{STL} and \texttt{CIFAR} for evaluation. Detailed information about these datasets are deferred to Appendix~\ref{appx-dataset}. Since the properties of these datasets vary case by case, we consider different ways to split the observed data into multiple domains or construct out-of-distribution data with new unseen domains (where the data from different domains are considered as samples from different distributions). We next describe the evaluation protocol in terms of how we split the training and testing sets with distribution shifts in each case. 

\textbf{$\circ$}\; \texttt{Twitch} is a multi-graph dataset~\cite{rozemberczki2021twitch} where nodes of graphs are instances whose interconnectivity is reflected by observed structures. Each subgraph is comprised of a social network of users from a particular region, so one can use the nodes in different subgraphs as samples from different domains, since these subgraphs have obviously different topological features~\cite{eerm}. In particular, we use the nodes from three subgraphs as training data (where we hold out $25\%$ for validation), and the nodes from the other three subgraphs as testing data. The task is a binary classification for users' genders, and the performance is measured by ROC-AUC.

\textbf{$\circ$}\; \texttt{Arxiv} is a temporal network~\cite{ogb-nips20} which evolves with time. Each node (i.e., a paper) as an instance has a time label indicating its publication year. Since the citation relationships among papers can significantly vary with different time-sensitive contexts, the distribution shifts naturally exist for the network data collected within different time windows. We thus use the papers published before 2014 as training data (where we hold out $25\%$ for validation), and the papers published after 2014 as testing data. The task is to predict the subareas of papers, and the performance is measured by Accuracy.

\textbf{$\circ$}\; \texttt{DPPIN} consists of multiple datasets of biological protein interactions~\cite{fu2022dppin}. Each dataset is comprised of protein instances corresponding to a particular protein identification method (that can be treated as the domain). Each protein has a time-evolving scalar representing the gene expression value and the proteins within each domain have certain interactions exposed by the co-expressed activities. The interaction networks at one time, reflecting the partially observed data geometries, can be noisy and incomplete given the dynamical evolution of the co-expression levels. We use the proteins of four datasets for training, one dataset for validation and the other seven datasets for testing. We consider the regression task for protein's gene expression values, and the performance is measured by RMSE.

\textbf{$\circ$}\; \texttt{STL-10} is an image dataset where each instance is an image. Since there are no observed structures interconnecting the instances, we create the inter-instance relations via $k$-nearest-neighbor ($k$NN) with the Euclidean distance between two instances' input features. We consider using different $k$'s for constructing the data from different domains. To be specific, the value of $k$ is set as $\{2, 3, 4, 8, 9, 10\}$ which results in six domains where each domain corresponds to a particular $k$NN graph interconnecting the fix set of instances. We split all the image instances into training and testing sets with the ratio 1:1. Then we use the training instances with the structures of the first three domains as training data (where we hold out $25\%$ for validation), and the testing instances with the structures of the other three domains as testing data. In this way, the training and testing data can be seen as samples with disparate underlying geometries. The task is to predict the image labels, and the performance is measured by Accuracy.

\textbf{$\circ$}\; \texttt{CIFAR-10} is another image dataset. Similarly, we use $k$NN to construct the inter-instance relations. Differently, we fix $k=5$ and consider cosine similarity as the distance function. To introduce distribution shifts, we add angle bias to the cosine similarity to construct data from different domains. To be specific, we denote by $s_i$ the distance function used for $k$NN in the $i$-th domain, and particularly we consider $s_i(\mathbf x_1, \mathbf x_2) = 1 - \mbox{cos}(\angle(\mathbf x_1, \mathbf x_2) + \theta_i)$ where $\angle$ denotes the angle of input feature vectors $\mathbf x_1$ and $\mathbf x_2$. We set $\theta_i \in \{ 0^o, 30^o, 90^o, 150^o, 160^o, 170^o\}$ , resulting in six domains. We adopt the same way for data splitting as \texttt{STL}, resulting in training data from three domains and testing data from the other three domains. The angle bias causes the interconnecting structures of training and testing data to have different proximity patterns.

\textbf{Competitors.} We compare with a set of learning algorithms including empirical risk minimization (ERM), which is the most commonly used supervised learning objective, and state-of-the-art approaches for improving generalization such as the adversarial learning approach DANN~\cite{DAGNN}, the distributionally robust optimization GroupDRO~\cite{dro}, the data augmentation approach Mixup~\cite{zhang2017mixup}, and invariant risk minimization approaches IRM~\cite{irm} and EERM~\cite{eerm}. These models are agnostic to encoder architectures. For fair comparison, we use GCN~\cite{GCN-vallina}, GAT~\cite{GAT} and DIFFormer~\cite{wu2023difformer} as their encoder backbones to compare with our models \modelgcn, \modelgat and \modeltrans, respectively. Due to space limit, we defer implementation details and hyper-parameter searching space to Appendix~\ref{appx-implementation}.

\begin{table*}[t!]
	\centering
	\caption{Testing Accuracy (\%) for \texttt{CIFAR} and \texttt{STL} on out-of-distribution data with different domains. We use the $k$-nearest-neighbor to construct the inter-instance relations in each domain. For \texttt{STL} (resp. \texttt{CIFAR}), different $k$'s (resp. angle bias for computing the cosine similarity) are used for creating distinct domains.\label{tbl-res-unobserved}}
	\begin{threeparttable}
        \setlength{\tabcolsep}{3mm}{ 
        \resizebox{\textwidth}{!}{
		\begin{tabular}{c|ccc|ccc}
		    \hline
		   \multirow{2}{*}{\textbf{Method}} & \multicolumn{3}{c|}{\texttt{CIFAR}} & \multicolumn{3}{c}{\texttt{STL}} \\ 
     
            & \textbf{$150^o$} & \textbf{$160^o$} & \textbf{$170^o$} & \textbf{$k=8$} & \textbf{$k=9$} & \textbf{$k=10$} \\  
 			\hline
        ERM-GCN  & 72.30 $\pm$ 0.22 & 73.16 $\pm$ 0.05 & 72.55 $\pm$ 0.30 & 72.89 $\pm$ 0.34 & 73.27 $\pm$ 0.40 & 74.18 $\pm$ 0.23 \\
        IRM-GCN & \textcolor{second}{\textbf{72.96 $\pm$ 0.32}} & \textcolor{second}{\textbf{73.97 $\pm$ 0.44}} & \textcolor{second}{\textbf{73.33 $\pm$ 0.08}} & \textcolor{third}{\textbf{72.96 $\pm$ 0.38}} & \textcolor{third}{\textbf{73.67 $\pm$ 0.22}} & 74.28 $\pm$ 0.40\\
    GroupDRO-GCN & \textcolor{third}{\textbf{72.73 $\pm$ 0.45}} & 73.40 $\pm$ 0.37 & \textcolor{third}{\textbf{72.85 $\pm$ 0.24}} & 72.79 $\pm$ 0.39 & 73.31 $\pm$ 0.19 & \textcolor{third}{\textbf{74.44 $\pm$ 0.67}}\\
    DANN-GCN   & 72.19 $\pm$ 0.34 & 73.12 $\pm$ 0.19 & 72.34 $\pm$ 0.37 & 72.77 $\pm$ 0.22 & 73.48 $\pm$ 0.56 & 74.38 $\pm$ 0.73 \\
    Mixup-GCN   & 72.66 $\pm$ 0.51 & \textcolor{third}{\textbf{73.49 $\pm$ 0.38}} & 72.60 $\pm$ 0.49 & \textcolor{second}{\textbf{73.04 $\pm$ 0.09}} & \textcolor{second}{\textbf{73.99 $\pm$ 0.19}} & \textcolor{second}{\textbf{74.68 $\pm$ 0.57}} \\
  EERM-GCN & 71.01 $\pm$ 0.95 & 72.29 $\pm$ 1.35 & 71.16 $\pm$ 1.36 & 72.14 $\pm$ 1.30& 72.11 $\pm$ 1.73 & 72.07 $\pm$ 1.41\\
  \textbf{\modelgcn} & \textcolor{first}{\textbf{79.24 $\pm$ 0.53}} & \textcolor{first}{\textbf{80.26 $\pm$ 0.60}} & \textcolor{first}{\textbf{79.39 $\pm$ 0.89}} & \textcolor{first}{\textbf{78.00 $\pm$ 0.34}} & \textcolor{first}{\textbf{78.63 $\pm$ 0.29}} & \textcolor{first}{\textbf{78.17 $\pm$ 0.29}}\\
            \hline
ERM-GAT & 72.92 $\pm$ 0.22 & \textcolor{third}{\textbf{74.39 $\pm$ 0.18}} & 73.22 $\pm$ 0.14 & 73.29 $\pm$ 0.36 & 73.38 $\pm$ 0.58 & 74.15 $\pm$ 0.74\\
IRM-GAT &  \textcolor{third}{\textbf{72.96 $\pm$ 0.17}} & 73.97 $\pm$ 0.38 & \textcolor{third}{\textbf{73.33 $\pm$ 0.10}} & 72.30 $\pm$ 1.42 & 73.16 $\pm$ 0.63 & 74.22 $\pm$ 1.24\\
  GroupDRO-GAT & 72.87 $\pm$ 0.18 & 74.07 $\pm$ 0.09 & 73.21 $\pm$ 0.11 & \textcolor{third}{\textbf{73.38 $\pm$ 0.19}} & \textcolor{second}{\textbf{73.70$\pm$ 0.12}} & 74.61 $\pm$ 0.51\\
  DANN-GAT & 72.81 $\pm$ 0.38 & 74.16 $\pm$ 0.15 & 73.23 $\pm$ 0.45 & \textcolor{second}{\textbf{73.45 $\pm$ 0.40}} & \textcolor{third}{\textbf{73.70 $\pm$ 0.19}} & \textcolor{second}{\textbf{74.90 $\pm$ 0.54}}\\
  Mixup-GAT & \textcolor{second}{\textbf{72.98 $\pm$ 0.17}} & \textcolor{second}{\textbf{74.41 $\pm$ 0.19}} & \textcolor{second}{\textbf{73.65 $\pm$ 0.13}} & 73.36 $\pm$ 0.21 & \textcolor{second}{\textbf{74.28 $\pm$ 0.17}} & \textcolor{third}{\textbf{74.75 $\pm$ 0.01}} \\
  EERM-GAT & 71.03 $\pm$ 1.69 & 72.82 $\pm$ 1.87 & 71.84 $\pm$ 1.14 & 71.31 $\pm$ 1.01 & 72.65 $\pm$ 1.89 & 71.96 $\pm$ 1.55\\
  \textbf{\modelgat} & \textcolor{first}{\textbf{78.23 $\pm$ 0.80}} & \textcolor{first}{\textbf{79.01 $\pm$ 0.63}} & \textcolor{first}{\textbf{77.98 $\pm$ 0.88}} & \textcolor{first}{\textbf{77.91 $\pm$ 0.84}} & \textcolor{first}{\textbf{79.17 $\pm$ 0.23}} & \textcolor{first}{\textbf{78.81 $\pm$ 0.66}}\\
  \hline
  ERM-Trans & 76.88 $\pm$ 0.11 & 77.51 $\pm$ 0.25 & 76.35 $\pm$ 0.28 & 76.53 $\pm$ 0.25 & 77.10 $\pm$ 0.65 & 77.90 $\pm$ 0.22 \\
  IRM-Trans & 76.53 $\pm$ 0.03 & 77.11 $\pm$ 0.05 & 76.42 $\pm$ 0.31 & 76.95 $\pm$ 0.14 & 77.49 $\pm$ 0.25 & 78.02 $\pm$ 0.35 \\
  GroupDRO-Trans & 76.94 $\pm$ 0.65 & 76.99 $\pm$ 0.31 & 76.37 $\pm$ 0.53 & \textcolor{third}{\textbf{77.81 $\pm$ 0.59}} & 78.01 $\pm$ 0.54 & 78.10 $\pm$ 0.27 \\
  DANN-Trans & 76.91 $\pm$ 0.17 & 77.13 $\pm$ 0.37 & 76.61 $\pm$ 0.30 & 77.64 $\pm$0.13 & 78.29 $\pm$ 0.54 & \textcolor{third}{\textbf{78.19 $\pm$ 0.35}}\\
  Mixup-Trans & \textcolor{third}{\textbf{77.49 $\pm$ 0.39}} & \textcolor{third}{\textbf{77.91 $\pm$ 0.14}} & \textcolor{third}{\textbf{77.45 $\pm$ 0.34}} & 77.76 $\pm$ 0.30 & \textcolor{third}{\textbf{78.32 $\pm$ 0.57}} & \textcolor{first}{\textbf{78.73 $\pm$ 0.76}} \\
  EERM-Trans & \textcolor{second}{\textbf{79.68 $\pm$ 0.51}} & \textcolor{second}{\textbf{79.89 $\pm$ 0.32}} & \textcolor{second}{\textbf{78.82 $\pm$ 0.54}} & \textcolor{second}{\textbf{77.92 $\pm$ 0.93}} & \textcolor{second}{\textbf{78.58 $\pm$ 0.20}} & 78.18 $\pm$ 0.38 \\
  \textbf{\modeltrans} & \textcolor{first}{\textbf{80.72 $\pm$ 0.39}} & \textcolor{first}{\textbf{81.06 $\pm$ 0.32}} & \textcolor{first}{\textbf{80.24 $\pm$ 0.38}} & \textcolor{first}{\textbf{78.06 $\pm$ 0.46}} & \textcolor{first}{\textbf{79.39 $\pm$ 0.28}} & \textcolor{second}{\textbf{78.41 $\pm$ 0.57}}\\
  \hline
		\end{tabular}}}
	\end{threeparttable}
\end{table*}

\subsection{Comparative Results}

\textbf{Generalization with Observed Geometries.} We first test the models on data with observed structures where we compare \modelgcn and \modelgat with the competitors using GCN and GAT as the encoder backbones, respectively. Table~\ref{tbl-res-observed} reports the performance on testing data of \texttt{Arxiv} and \texttt{Twitch}. In particular, for \texttt{Arxiv}, we further divide the testing data into three subsets according to the publication years of the papers to closely check the generalization performance on testing data with different levels of distribution shifts. As we can see, the accuracy produced by all methods exhibits an overall decrease as the time gap (between training and testing data) enlarges, while \modelgcn and \modelgat significantly alleviate the performance drop. Notably, for testing data within 2018-2020, \modelgcn and \modelgat achieve $23.8\%$ and $26.6\%$ improvement of the accuracy over the runner-ups, respectively. Furthermore, for \texttt{Twitch}, we separately report the ROC-AUC on three testing subgraphs to compare the generalization performance on specific cases. The results show that \modelgcn and \modelgat achieve overall superior ROC-AUCs over these strong competitors, respectively.

\textbf{Generalization with Partially Observed Geometries.} We next consider the case where the data geometries are partially observed. This situation can be commonly encountered in practice, as the observed relational structures can be noisy and incomplete. To accommodate the potential interactions, we use our model \modeltrans and compare it with the competitors using DIFFormer as the encoder. Table~\ref{tbl-res-dppin} reports the performance on testing data of \texttt{DPPIN}. We found that our model \modeltrans consistently achieves the first-ranking results (the lowest RMSE) in seven testing domains, with an averaged reduction of $21.6\%$ on RMSE over the runner-up. This demonstrates the practical efficacy of the proposed model for handling generalization with partially observed data geometries.

\textbf{Generalization with Unobserved Geometries.} The last scenario we study involves data geometries that are unobserved. We use the image datasets for demonstration where the inter-instance structures are synthetically constructed through the $k$-nearest-neighbor. We adopt the three model versions for comparison, where \modelgcn and \modelgat purely rely on the input structures and \modeltrans estimates the potential interactions among arbitrary instance pairs. Table~\ref{tbl-res-unobserved} reports the performance on testing data of \texttt{CIFAR} and \texttt{STL}. In contrast with the competitors, our model yields the highest averaged accuracy on two datasets, demonstrating the superiority of the proposed model. Comparing the models using different backbones, we found that the Transformer-style architecture contributes to more stable performance (with lower variance across different testing sets). This is probably because the Transformer-style architecture can capture the latent interactions that are more insensitive to distribution shifts of structures than the GNNs that purely rely on observed structures. 

\subsection{Further Discussions}

\begin{figure}[!h]
\centering
    \begin{minipage}{\linewidth}
    \centering
   \subfigure[Accuracy on testing sets]{
   \label{fig-abaltion}
    \begin{minipage}[t]{0.45\linewidth}
    \centering
    \includegraphics[width=0.99\linewidth]{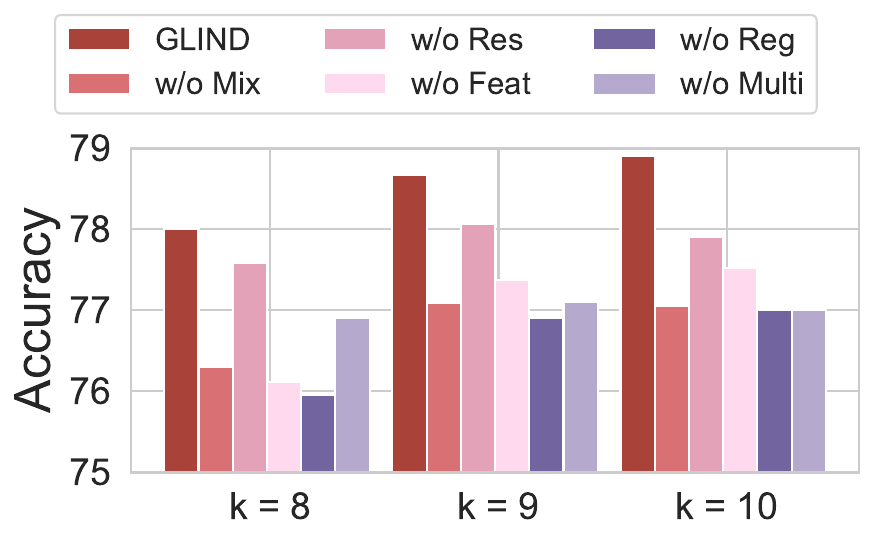}
    \end{minipage}
    }
    \subfigure[Accuracy on testing sets]{
    \label{fig-hyper}
    \begin{minipage}[t]{0.46\linewidth}
    \centering
    \includegraphics[width=0.95\linewidth]{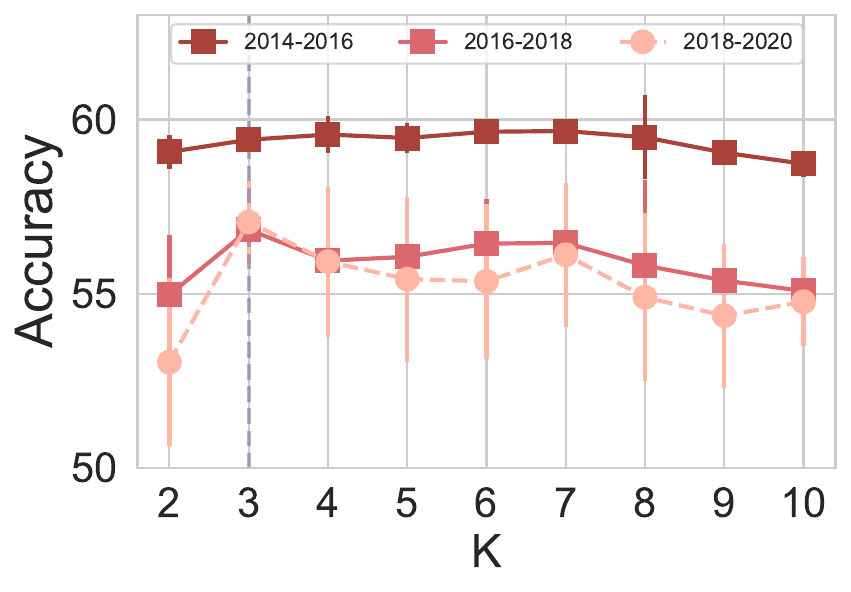}
    \end{minipage}
    }
    \label{fig-further-study}
    \caption{(a) Ablation studies for \model on \texttt{STL}. (b) Performance of \model on \texttt{Arxiv} with different $K$'s.}
    \end{minipage}
\end{figure}

\textbf{Ablation Studies.} To further verify the effectiveness of \model, we next dissect the efficacy of the specific proposed components in our model. We conduct ablation studies w.r.t. model architectures and learning objectives, respectively, and report the comparison results on \texttt{STL} in Fig.~\ref{fig-abaltion} (more results are deferred to Appendix~\ref{appx-res}). For the model architecture, we compare with three variants: 1) \emph{w/o Res} removes the residual link ($\mathbf z_u^{(l)}$ added to each layer); 2) \emph{w/o Feat} removes the self-loop feature aggregation ($\mathbf W_S^{(l, k)} \mathbf z_u^{(l)}$); 3) \emph{w/o Multi} sets $K=1$ that removes the multiple branches at each layer. We found that all of these components contribute to consistent performance improvements in the three testing domains, which verifies the efficacy of these designs and our model architecture induced by the principled diffusion framework. For the learning objective, we compare with two variants: 1) \emph{w/o Reg} removes the regularization term (the KL loss in Eqn.~\ref{eqn-obj-reweight2}); 2) \emph{w/o Mix} replaces the prior distribution in Eqn.~\ref{eqn-prior} with the average of variational posterior $p_0(\mathbf d^{(l)}) \approx 1/N \sum_{u=1}^N q(\mathbf d^{(l)}|\mathbf z^{(l)}=\mathbf z^{(l)}_u)$. We found that these two variants lead to some performance drops, which verifies the effectiveness of our learning objective. 

\textbf{Impact of Numbers of Diffusivity Hypothesis.} We proceed to study the impact of the hyper-parameter $K$, i.e., the number of diffusivity hypothesis at each message passing layer, on the model performance. Fig.~\ref{fig-hyper} presents the testing accuracy of \modelgcn on \texttt{Arxiv} with $K$ ranging from 2 to 10 (more results are deferred to Appendix~\ref{appx-res}). We found that setting $K$ as a moderate value can bring up overall the optimal performance. If $K$ is too small, the model would have insufficient capacity for learning multi-faceted data interactions from complex underlying generative mechanisms; if $K$ is too large, the model would become overly complicated which might impair generalization.

\section{Conclusion}\label{sec-conclusion}

This paper has explored a geometric diffusion framework for generalization with interdependent data. The model is generalized from the diffusion process on manifolds with stochastic diffusivity function generated from a variational distribution, conditioned on the diffusion dynamics trajectories. We further develop a re-weighting regularization approach to guide the diffusion model to learn causal relations from inputs to outputs throughout the diffusion process that facilitates generalization. Experiments on diverse real-world datasets with different kinds of distribution shifts verify the effectiveness of the proposed models.

\textbf{Future Work.} There can be several promising directions for future investigation. For example, apart from the experimental scenarios and datasets considered in this paper, the proposed methodology and general ideas can be potentially applied and extended to other tasks where out-of-distribution generalization matters, e.g., molecule representations~\cite{moleood}. Another problem closely related to out-of-distribution generalization is out-of-distribution detection~\cite{gnnsafe,graphde}, whose goal is to enable the model to identify testing data from different distributions than training. It also remains under-explored how the diffusion framework explored in this paper can be used for guiding the methodology in the context of out-of-distribution detection with interdependent data. 

\section*{Impact Statement}

This work presents an attempt to improve the generalization for handling interdependent data. Researching the generalization of machine learning models holds profound implications for advancing the reliability and applicability of artificial intelligence systems. Understanding and enhancing the generalization capabilities of models is paramount for deploying them across diverse real-world scenarios. Improved generalization not only enhances model performance on unseen data but also addresses ethical concerns related to bias and fairness. Furthermore, research in this domain contributes to the creation of models that are transferable across different domains, fostering broader adoption of machine learning solutions. As the impact of AI continues to grow across various sectors, unraveling the intricacies of generalization becomes instrumental in ensuring the responsible and equitable deployment of machine learning technologies.

\section*{Acknowledgement}

The work was in part supported by NSFC (92370201, 62222607, 72342023).

\bibliographystyle{icml2024}
\bibliography{main}

\newpage

\newpage
\appendix
\onecolumn
\section*{Appendix}

\section{Derivation and Proof}\label{appx-prove}


\subsection{Derivation for Eqn.~\ref{eqn-obj-var2}}\label{appx-prove-1}

The log-likelihood for observed data $\log p_\theta(\mathbf y|\mathbf x, \mathcal G)$ can be computed by
\begin{equation}\label{eqn-proof-obj-1}
\begin{split}
\log p_\theta(\mathbf y|\mathbf x, \mathcal G) & =
    \log \sum_{\mathbf z^{(1)}, \cdots, \mathbf z^{(L-1)}}\prod_{l=0}^{L-1} p_\theta(\mathbf z^{(l+1)} | \mathbf z^{(l)}, \mathcal G) \\
    & =
    \sum_{l=0}^{L-2} \log \sum_{\mathbf z^{(l+1)}} p_\theta(\mathbf z^{(l+1)} | \mathbf z^{(l)}, \mathcal G) + \log p_\theta (\mathbf z^{(L)} | \mathbf z^{(L-1)}, \mathcal G) \\
    & =  \sum_{l=0}^{L-2} \log \sum_{\mathbf z^{(l+1)}}\mathbb E_{p(\mathbf d^{(l)} | \mathbf z^{(l)})} [ p_\theta(\mathbf z^{(l+1)} | \mathbf z^{(l)}, \mathbf d^{(l)}, \mathcal G) ]\\
    & + \log \mathbb E_{p(\mathbf d^{(L-1)} | \mathbf z^{(L-1)})} [ p_\theta(\mathbf z^{(L)} | \mathbf z^{(L-1)}, \mathbf d^{(L-1)}, \mathcal G) ],
\end{split}
\end{equation}
where the marginalization over $\mathbf z^{(1)}, \cdots, \mathbf z^{(L-1)}$ (which is omitted in Eqn.~\ref{eqn-obj-ori} for brevity) is through the delta predictive distributions $p_\theta(\mathbf z^{(l+1)} | \mathbf z^{(l)}, \mathbf d^{(l)}, \mathcal G)$. Due to the intractable integration over $\mathbf d^{(l)}$, we consider the evidence lower bound of Eqn.~\ref{eqn-proof-obj-1} by introducing $q_\phi(\mathbf d^{(l)} | \mathbf z^{(l)})$:
\begin{equation}\label{eqn-proof-obj-2}
\begin{split}
    & \sum_{l=0}^{L-2} \mathbb E_{q_\phi(\mathbf d^{(l)} | \mathbf z^{(l)} )} \left [ \log \sum_{\mathbf z^{(l+1)}} p_\theta(\mathbf z^{(l+1)} | \mathbf z^{(l)}, \mathbf d^{(l)}, \mathcal G) \frac{p(\mathbf d^{(l)} | \mathbf z^{(l)} )}{q_\phi(\mathbf d^{(l)} | \mathbf z^{(l)} )} \right ] \\
    + & \mathbb E_{q_\phi(\mathbf d^{(L-1)} | \mathbf z^{(L-1)} )} \left [ \log p_\theta(\mathbf z^{(L)} | \mathbf z^{(L-1)}, \mathbf d^{(L-1)}, \mathcal G) \frac{p(\mathbf d^{(L-1)} | \mathbf z^{(L-1)} )}{q_\phi(\mathbf d^{(L-1)} | \mathbf z^{(L-1)} )} \right ] \\
    = & \sum_{l=0}^{L-2}\mathbb E_{q_\phi(\mathbf d^{(0)} | \mathbf z^{(0)}), \cdots, q_\phi(\mathbf d^{(L-2)} | \mathbf z^{(L-2)} )} [
    \log \sum_{\mathbf z^{(l+1)}} p_\theta(\mathbf z^{(l+1)} | \mathbf z^{(l)}, \mathbf d^{(l)}, \mathcal G)
    ] \\
    + & \mathbb E_{q_\phi(\mathbf d^{(L-1)} | \mathbf z^{(L-1)})} [\log p_\theta(\mathbf z^{(L)} | \mathbf z^{(L-1)}, \mathbf d^{(L-1)}, \mathcal G)] + \sum_{l=0}^{L-1} \mathbb E_{q_\phi(\mathbf d^{(l)} | \mathbf z^{(l)} )} \left [ \log  \frac{p(\mathbf d^{(l)} | \mathbf z^{(l)} )}{q_\phi(\mathbf d^{(l)} | \mathbf z^{(l)} )} \right ] \\
    = & \mathbb E_{q_\phi(\mathbf d^{(0)} | \mathbf z^{(0)}), \cdots, q_\phi(\mathbf d^{(L-1)} | \mathbf z^{(L-1)} )} [\log p_\theta (\mathbf z^{(L)}|\mathbf z^{(0)}, \mathbf d^{(0)}, \cdots, \mathbf d^{(L-1)}, \mathcal G)] \\
    - & \sum_{l=0}^{L-1}\mbox{KL}(q_\phi(\mathbf d^{(l)} | \mathbf z^{(l)} ), p(\mathbf d^{(l)} | \mathbf z^{(l)} )).
\end{split}
\end{equation}
This gives rise to the objective of Eqn.~\ref{eqn-obj-var2}. 

\subsection{Proof for Theorem~\ref{thm-1}}\label{appx-prove-2}

Before the proof, we first introduce two fundamental rules of $do$-calculus~\cite{causal-old1} which will be used as the building blocks later. Consider a causal directed acyclic graph $\mathcal A$ with three nodes: $B$, $D$ and $E$. We denote $\mathcal A_{\overline{B}}$ as the intervened causal graph by cutting off all arrows coming into $B$, and $\mathcal A_{\underline{B}}$ as the graph by cutting off all arrows going out from $B$. For any interventional distribution compatible with $\mathcal A$, the $do$-calculus induces the following two fundamental rules. 

    i) Action/observation exchange:
\begin{equation}
    P(d|do(b),do(e)) = P(d|do(b),e),\;\text{if}\;(D \perp \!\!\! \perp E|B)_{\mathcal A_{\overline{B}\underline{E}}}. \nonumber
\end{equation}

    ii) Insertion/deletion of actions:
\begin{equation}
    P(d|do(b),do(e)) = P(d|do(b)),\;\text{if}\;(D\perp \!\!\! \perp E|B)_{\mathcal A_{\overline{BE}}}. \nonumber
\end{equation}

Back to our case where we study the causal graph comprised of $\mathbf x$, $\mathbf y$ and $\mathbf d$. We have 
\begin{equation}\label{eqn-proof2-1}
\begin{split}
    p_\theta(\mathbf y|do(\mathbf x), \mathcal G) & = \sum_{\mathbf d}  p_\theta(\mathbf y|do(\mathbf x), \mathbf d, \mathcal G) p(\mathbf d|do(\mathbf x)) \\
    & = \sum_{\mathbf d}  p_\theta(\mathbf y|\mathbf x, \mathbf d, \mathcal G) p(\mathbf d|do(\mathbf x)) \\
    & = \sum_{\mathbf d}  p_\theta(\mathbf y|\mathbf x, \mathbf d, \mathcal G) p_0(\mathbf d),
\end{split}
\end{equation}
where the first step is given by the law of total probability, the second step is according to the first rule (since $\mathbf y\perp \!\!\! \perp \mathbf x |\mathbf d$ in $\mathcal A_{\underline {\mathbf x}}$), and the third step is due to the second rule (since we have $\mathbf d \perp \!\!\! \perp \mathbf x$ in $\mathcal A_{\overline {\mathbf x}}$). The above derivation shows that $p_\theta(\mathbf y|do(\mathbf x), \mathcal G) = \mathbb E_{p_0(\mathbf d)}[p_\theta (\mathbf y|\mathbf x, \mathbf d, \mathcal G)]$ where $p_0$ is the prior distribution of the diffusivity.

Now consider the diffusion model that induces a trajectory $\mathbf x = \mathbf z^{(0)} \rightarrow \mathbf z^{(1)} \rightarrow \cdots \rightarrow \mathbf z^{(L)} = \mathbf y$. By treating $\{\mathbf d^{(l)}\}_{l=0}^{L-1}$ as a whole random variable, we can extend the result of Eqn.~\ref{eqn-proof2-1} and obtain
\begin{equation}
    p_\theta(\mathbf y|do(\mathbf x), \mathcal G) = \sum_{\mathbf d^{(0)}, \cdots, \mathbf d^{(L-1)}} p_\theta(\mathbf y|\mathbf x, \mathbf d^{(0)}, \cdots, \mathbf d^{(L-1)}, \mathcal G) p_0(\mathbf d^{(0)}, \cdots, \mathbf d^{(L-1)}).
\end{equation}
Notice that $p_0(\mathbf d^{(0)}, \cdots, \mathbf d^{(L-1)}) = \prod_{l=0}^{L-1} p_0(\mathbf d^{(l)})$. By inserting the delta predictive distribution $p_\theta(\mathbf z^{(l+1)}| \mathbf z^{(l)}, \mathbf d^{(l)}, \mathcal G)$ induced by the diffusion model, we can derive the log-likelihood: 
\begin{equation}\label{eqn-proof2-2}
\begin{split}
    & \log p_\theta(\mathbf y|do(\mathbf x), \mathcal G) \\
    = & \log \sum_{\mathbf z^{(1)}, \cdots, \mathbf z^{(L-1)}}\sum_{\mathbf d^{(0)}, \cdots, \mathbf d^{(L-1)}} \prod_{l=0}^{L-1} p_\theta(\mathbf z^{(l+1)}| \mathbf z^{(l)}, \mathbf d^{(l)}, \mathcal G) p_0(\mathbf d^{(l)}) \\
    = & \sum_{l=0}^{L-2} \log \sum_{\mathbf z^{(l+1)}}\sum_{\mathbf d^{(l)}} p_\theta(\mathbf z^{(l+1)}| \mathbf z^{(l)}, \mathbf d^{(l)}, \mathcal G) p_0(\mathbf d^{(l)}) + \log \sum_{\mathbf d^{(L-1)}} p_\theta(\mathbf z^{(L)}| \mathbf z^{(L-1)}, \mathbf d^{(L-1)}, \mathcal G) p_0(\mathbf d^{(L-1)}) \\
    = & \sum_{l=0}^{L-2} \log \sum_{\mathbf z^{(l+1)}}\sum_{\mathbf d^{(l)}} p_\theta(\mathbf z^{(l+1)}| \mathbf z^{(l)}, \mathbf d^{(l)}, \mathcal G) p_0(\mathbf d^{(l)}) + \log \sum_{\mathbf d^{(L-1)}} p_\theta(\mathbf z^{(L)}| \mathbf z^{(L-1)}, \mathbf d^{(L-1)}, \mathcal G) p_0(\mathbf d^{(L-1)}) \\
\end{split}
\end{equation}
We next separately consider the derivation for the two terms in the above equation. For the first term in Eqn.~\ref{eqn-proof2-2}, using Jensen Inequality we have
\begin{equation}\label{eqn-proof2-3}
    \begin{split}
        & \sum_{l=0}^{L-2} \log \sum_{\mathbf z^{(l+1)}}\sum_{\mathbf d^{(l)}} p_\theta(\mathbf z^{(l+1)}| \mathbf z^{(l)}, \mathbf d^{(l)}, \mathcal G) p_0(\mathbf d^{(l)}) \\
        = & \sum_{l=0}^{L-2} \log \sum_{\mathbf z^{(l+1)}}\sum_{\mathbf d^{(l)}} p_\theta(\mathbf z^{(l+1)}| \mathbf z^{(l)}, \mathbf d^{(l)}, \mathcal G) p_0(\mathbf d^{(l)})\frac{q_\phi(\mathbf d^{(l)}|\mathbf z^{(l)})}{q_\phi(\mathbf d^{(l)}|\mathbf z^{(l)})} \\
        \geq & \sum_{l=0}^{L-2} \sum_{\mathbf d^{(l)}} q_\phi(\mathbf d^{(l)}|\mathbf z^{(l)}) 
        \cdot \left [\log \sum_{\mathbf z^{(l+1)}} p_\theta(\mathbf z^{(l+1)}| \mathbf z^{(l)}, \mathbf d^{(l)}, \mathcal G) \frac{p_0(\mathbf d^{(l)})}{q_\phi(\mathbf d^{(l)}|\mathbf z^{(l)})} \right ].
    \end{split}
\end{equation}
For the second term in Eqn.~\ref{eqn-proof2-2}, similarly we have
\begin{equation}\label{eqn-proof2-4}
    \begin{split}
        & \log \sum_{\mathbf d^{(L-1)}} p_\theta(\mathbf z^{(L)}| \mathbf z^{(L-1)}, \mathbf d^{(L-1)}, \mathcal G) p_0(\mathbf d^{(L-1)}) \\
        \geq & \sum_{\mathbf d^{(L-1)}} q_\phi(\mathbf d^{(L-1)}|\mathbf z^{(L-1)}) 
        \cdot \left [\log p_\theta(\mathbf z^{(L)}| \mathbf z^{(L-1)}, \mathbf d^{(L-1)}, \mathcal G) \frac{p_0(\mathbf d^{(L-1)})}{q_\phi(\mathbf d^{(L-1)}|\mathbf z^{(L-1)})} \right ].
    \end{split}
\end{equation}
By combing the results of Eqn.~\ref{eqn-proof2-3} and Eqn.~\ref{eqn-proof2-4} with Eqn.~\ref{eqn-proof2-2}, we have the variational lower bound of $\log p_\theta(\mathbf y|do(\mathbf x), \mathcal G)$:
\begin{equation}\label{eqn-proof2-5}
\begin{split}
    & \sum_{l=0}^{L-2} \mathbb E_{q_\phi(\mathbf d^{(l)} | \mathbf z^{(l)} )} \left [ \log \sum_{\mathbf z^{(l+1)}} p_\theta(\mathbf z^{(l+1)} | \mathbf z^{(l)}, \mathbf d^{(l)}, \mathcal G) \frac{p_0(\mathbf d^{(l)} )}{q_\phi(\mathbf d^{(l)} | \mathbf z^{(l)} )} \right ] \\
    + & \mathbb E_{q_\phi(\mathbf d^{(L-1)} | \mathbf z^{(L-1)} )} \left [ \log p_\theta(\mathbf z^{(L)} | \mathbf z^{(L-1)}, \mathbf d^{(L-1)}, \mathcal G) \frac{p_0(\mathbf d^{(L-1)} )}{q_\phi(\mathbf d^{(L-1)} | \mathbf z^{(L-1)} )} \right ],
\end{split}
\end{equation}
where the marginalization over $\mathbf z^{(1)}, \cdots, \mathbf z^{(L-1)}$ (which is omitted in Eqn.~\ref{eqn-obj-reweight} for brevity) is through the delta predictive distribution $p_\theta(\mathbf z^{(l+1)} | \mathbf z^{(l)}, \mathbf d^{(l)}, \mathcal G)$.

Furthermore, in Eqn.~\ref{eqn-proof2-3} the equality holds if and only if $q_\phi(\mathbf d^{(l)}|\mathbf z^{(l)}) = p(\mathbf d^{(l)} | \mathbf z^{(l+1)}, \mathbf z^{(l)}, \mathcal G) \cdot \frac{p_0(\mathbf d^{(l)})}{p(\mathbf d^{(l)} | \mathbf z^{(l)})}$ in which case the function inside the expectation stays as a constant w.r.t. $\mathbf d^{(l)}$. Similarly, for Eqn.~\ref{eqn-proof2-4}, the equality holds if and only if $q_\phi(\mathbf d^{(L-1)}|\mathbf z^{(L-1)}) = p(\mathbf d^{(L-1)} | \mathbf z^{(L)}, \mathbf z^{(L-1)}, \mathcal G)\cdot \frac{p_0(\mathbf d^{(L-1)})}{p(\mathbf d^{(L-1)} | \mathbf z^{(L-1)})}$.

\subsection{Derivation for Eqn.~\ref{eqn-obj-reweight2}}\label{appx-prove-3}

The objective Eqn.~\ref{eqn-obj-reweight2} can be derived in a similar way as Eqn.~\ref{eqn-proof-obj-2} on the basis of Eqn.~\ref{eqn-proof2-5}. In specific, we leverage the model-free prior $p_0(\mathbf d^{(l)})$ to replace $p(\mathbf d^{(l)} | \mathbf z^{(l)} )$ in the first line of Eqn.~\ref{eqn-proof-obj-2}:
\begin{equation}\label{eqn-proof-obj-3}
\begin{split}
     & \sum_{l=0}^{L-2} \mathbb E_{q_\phi(\mathbf d^{(l)} | \mathbf z^{(l)} )} \left [ \log \sum_{\mathbf z^{(l+1)}} p_\theta(\mathbf z^{(l+1)} | \mathbf z^{(l)}, \mathbf d^{(l)}, \mathcal G) \frac{p_0(\mathbf d^{(l)} )}{q_\phi(\mathbf d^{(l)} | \mathbf z^{(l)} )} \right ] \\
    + & \mathbb E_{q_\phi(\mathbf d^{(L-1)} | \mathbf z^{(L-1)} )} \left [ \log p_\theta(\mathbf z^{(L)} | \mathbf z^{(L-1)}, \mathbf d^{(L-1)}, \mathcal G) \frac{p_0(\mathbf d^{(L-1)} )}{q_\phi(\mathbf d^{(L-1)} | \mathbf z^{(L-1)} )} \right ] \\
    = & \mathbb E_{q_\phi(\mathbf d^{(0)} | \mathbf z^{(0)}), \cdots, q_\phi(\mathbf d^{(L-1)} | \mathbf z^{(L-1)} )} [\log p_\theta (\mathbf z^{(L)}|\mathbf z^{(0)}, \mathbf d^{(0)}, \cdots, \mathbf d^{(L-1)}, \mathcal G)] \\
    - & \sum_{l=0}^{L-1}\mbox{KL}(q_\phi(\mathbf d^{(l)} | \mathbf z^{(l)} ), p_0(\mathbf d^{(l)} )).
\end{split}
\end{equation}

\section{Dataset Introduction}\label{appx-dataset}



\texttt{Twitch}~\cite{rozemberczki2021twitch} is a social network dataset which comprises multiple disconnected subgraphs. Each subgraph is a social network from a particular region and contains thousands of node instances that are interconnected. Each node is a use on Twitch and the node label is the gender of the user. Edges denote the friendship among users. Since in this dataset, nodes from different subgraphs can be seen as samples from distinct distributions (Table~\ref{tbl-twitch-stat} shows different statistics of these subgraphs), we split the training and testing data according to subgraphs. Specifically, we use the nodes from subgraphs DE, PT and RU as training data, and the nodes from subgraphs ES, FR and EN as testing data. 

\begin{figure}[!h]
\centering
    \begin{minipage}{\linewidth}
    \centering
   \subfigure[2014-2016 test instances]{
    \begin{minipage}[t]{0.25\linewidth}
    \centering
    \includegraphics[width=0.8\linewidth]{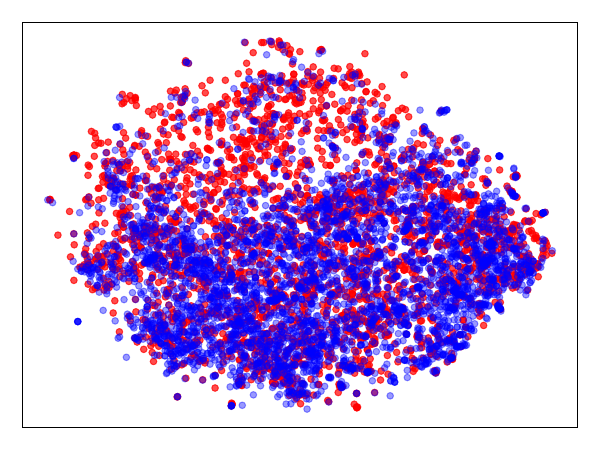}
    \end{minipage}
    }
    \subfigure[2016-2018 test instances]{
    \begin{minipage}[t]{0.25\linewidth}
    \centering
    \includegraphics[width=0.8\linewidth]{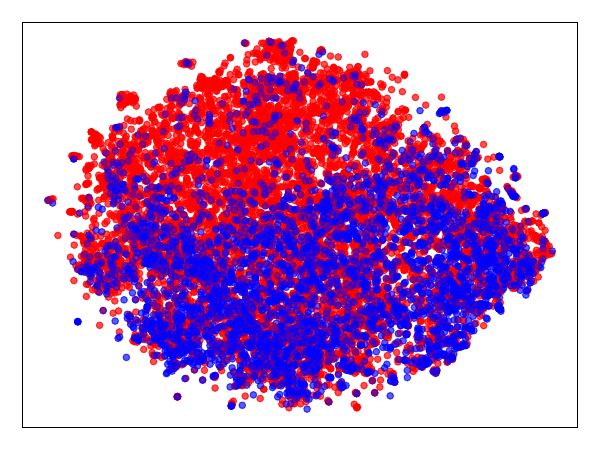}
    \end{minipage}
    }
    \subfigure[2018-2020 test instances]{
    \begin{minipage}[t]{0.25\linewidth}
    \centering
    \includegraphics[width=0.8\linewidth]{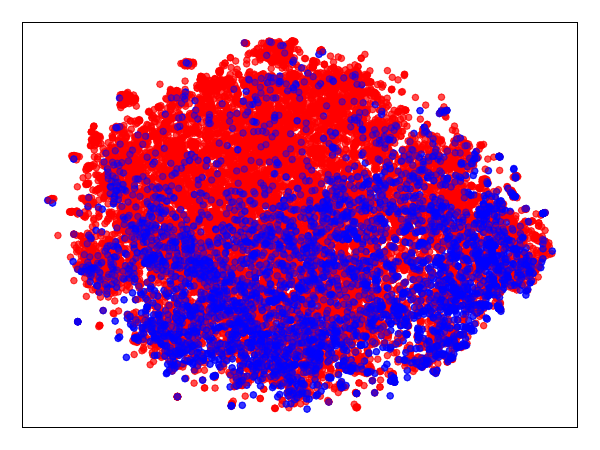}
    \end{minipage}
    }
    \caption{T-SNE visualization of input features for training instances (blue) and testing instances (red) on \texttt{Arxiv}.\label{tbl-arxiv-stat}}
    \end{minipage}
\end{figure}

\texttt{Arxiv} is a citation network dataset provided by \cite{OGB}. It records the citation relationship between papers and the goal is to predict the paper's subarea based on the word embeddings of paper keywords as the node features. The dataset also provides the time information about when the paper is published, and we use the publication year to split the training and testing sets. Since the research subareas and citation relationships of papers are sensitive to time, there natural exist distribution shifts between papers published in different time windows. In our experiments, we use papers published before 2014 for training, and papers published after 2014 for testing. We further divide the testing data into three subsets comprised of the papers published within 2014-2016, 2016-2018 and 2018-2020, respectively. As verification for the distribution shifts, Table~\ref{tbl-arxiv-stat} visualizes the different distributions of training and testing instances. 

\texttt{CIFAR-10} and \texttt{STL-10} are two image datasets where each image is an instance and there are no observed structures interconnecting the instances. Following the recent work \cite{wu2023difformer}, we use all the 13000 images, each of which belongs to one of ten classes, for \texttt{STL-10}, and choose 1500 images from each of 10 classes for \texttt{CIFAR-10}. The input features of image instances are pre-computed by using the self-supervised approach SimCLR~\citep{chen2020simple} (that does not use labels for training) and training a ResNet-18 that extracts the feature map as the input features of each instance. We use the $k$-nearest-neighbor to construct the inter-instance relational structures for these datasets, and consider different $k$'s and distance functions to introduce the distribution shifts. More details about how we choose the values of $k$ and distance functions are presented in Sec.~\ref{sec-exp-setup}.

\texttt{DPPIN}~\citep{fu2022dppin} is comprised of 12 individual dynamic network datasets at different scales, and each network records the protein-protein interactions. Each dynamic network is obtained by one protein identification method, which can be seen as one domain, and consists of 36 snapshots. Each protein has a sequence of continuous scalar features with 36 time stamps, which records the evolution of gene expression values within metabolic cycles of yeast cells. The interconnecting structures among protein instances are determined by co-expressed protein pairs at one time, and can change as the co-expression activities evolve with time. We consider the regression task for the gene expression value of each protein instance within one snapshot, and ignore the temporal dependence between different snapshots. In specific, the model input contains the proteins' gene expression value in the previous ten snapshot (as input features) and the interaction structures at the current snapshot, based on which the prediction target is the gene expression value of proteins at the current snapshot. As observed by \citep{fu2022dppin}, the protein instances from different datasets have distinct topological patterns (e.g., the distributions of cliques), so we split the training and testing data based on different datasets.

\begin{table}[t]
\centering
\caption{Statistics for different subgraphs of \texttt{Twitch}.\label{tbl-twitch-stat}}
    \resizebox{0.5\textwidth}{!}{
\begin{tabular}{@{}c|cccccc@{}}
\toprule
\textbf{Datasets} & \textbf{DE} & \textbf{PT} & \textbf{RU} & \textbf{ES}  & \textbf{FR}  & \textbf{EN}  \\ 
\midrule
\textbf{\# Instances} &  9498 & 1912 & 4385 & 4648 & 6549 & 7126 \\
\textbf{Density} &  0.0033 & 0.0171 & 0.0038 & 0.0054 & 0.0052 & 0.0013 \\
\textbf{Max Node Degree} &  3475 & 455 & 575 & 809 & 1517 & 465 \\
\textbf{Mean Node Degree} &  16 & 16 & 8 & 12 & 17 & 4 \\
\bottomrule
\end{tabular}}
\end{table}

\section{Implementation Details}\label{appx-implementation}

Our implementation is based on PyTorch 1.9.0 and PyTorch Geometric 2.0.3. All of our experiments are run on a Tesla V100 with 16 GB memory. We adopt Adam with weight decay for training. We set a fixed training budget with 1000 epochs for \texttt{DPPIN} and 500 epochs for other datasets. The testing performance achieved by the epoch where the model gives the best performance on validation data is reported.

The detailed architecture of \model is described as follows. The model architecture consists of the following modules in sequential order:

\noindent\textbf{$\bullet$\;} A fully-connected layer with hidden size $D\times d$ (transforming $D$-dim input raw features into $d$-dim embeddings).

\noindent\textbf{$\bullet$\;} $L$-layer message passing network with hidden size $d$ (each layer contains $K$ branches that have independent parameterization), based on the three instantiations in Sec.~\ref{sec-model}.

\noindent\textbf{$\bullet$\;} A fully-connected layer with hidden size $d\times C$ (mapping $d$-dim embeddings to $C$ classes).

In each layer, we use ReLU activation, dropout and the residual link where the weight $\alpha$ is set as 0.5 across all datasets. 

The probability estimation for each branch is modeled by a feed-forward layer with hidden size $d\times K$ in each layer as introduced in Sec.~\ref{sec-model}. The sampling process for the selected branch is approximated by Gumbel Softmax. For computing the prior distribution in the objective, we introduce a pseudo dataset of size $T$ as described in Eqn.~\ref{eqn-prior}. We set $T = \mbox{int}(0.01 * N)$, where $N$ denotes the number of instances in the dataset, for all the datasets.

For other hyper-parameters, we search them for each dataset with grid search on the validation set. The searching spaces for all the hyper-parameters are as follows.

\noindent\textbf{$\bullet$\;} Number of message passing layers $L$: [2, 3, 4, 5].

\noindent\textbf{$\bullet$\;} Hidden dimension $d$: [32, 64, 128].

\noindent\textbf{$\bullet$\;} Dropout ratio: [0.0, 0.1, 0.2, 0.5].

\noindent\textbf{$\bullet$\;} Learning rate: [0.001, 0.005, 0.01, 0.02].

\noindent\textbf{$\bullet$\;} Weight decay: [0, 5e-5, 5e-4, 1e-3].

\noindent\textbf{$\bullet$\;} Number of diffusivity hypothesis $K$: [3, 4, 5, 10].

\noindent\textbf{$\bullet$\;} Gumbel Softmax temperature $\tau$: [1, 2, 3, 5].

\section{Additional Experiment Results}\label{appx-res}

\begin{figure}[!h]
\vspace{-10pt}
\centering
    \begin{minipage}{\linewidth}
    \centering
   \subfigure[Accuracy on \texttt{CIFAR}]{
   \label{fig-abaltion}
    \begin{minipage}[t]{0.35\linewidth}
    \centering
    \includegraphics[width=0.99\linewidth]{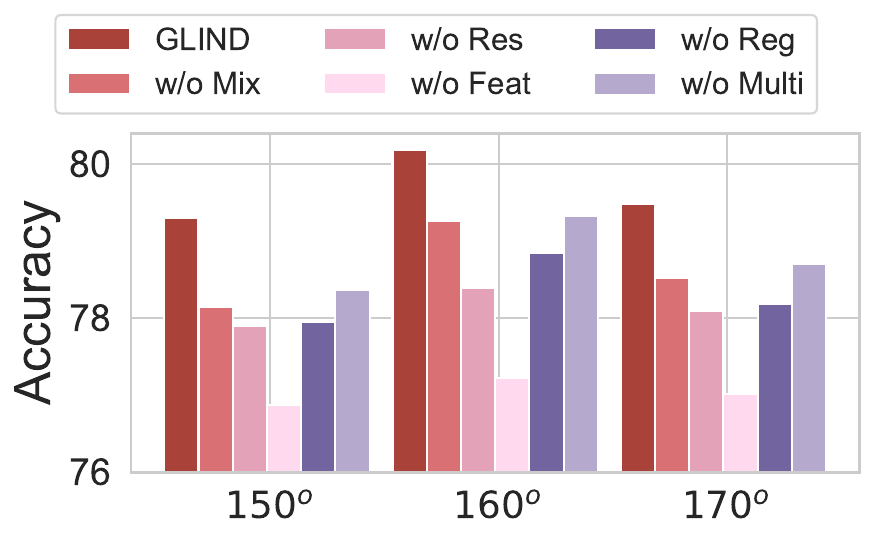}
    \end{minipage}
    }
    \subfigure[Accuracy on \texttt{Twitch}]{
    \label{fig-hyper}
    \begin{minipage}[t]{0.35\linewidth}
    \centering
    \includegraphics[width=0.95\linewidth]{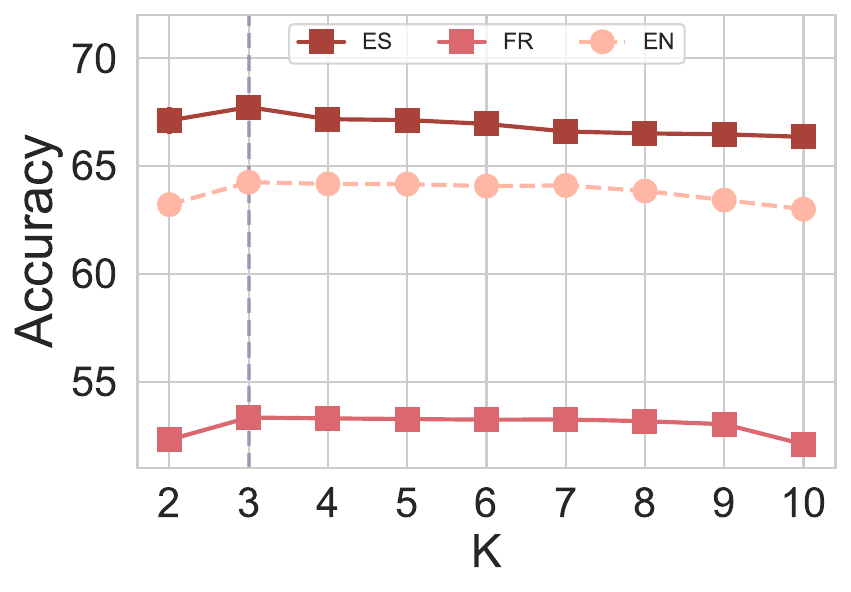}
    \end{minipage}
    }
    \vspace{-10pt}
    \label{fig-further-study}
    \caption{(a) Ablation studies for \model on \texttt{CIFAR}. (b) Performance of \model on three testing sets of \texttt{Twitch} with different numbers of diffusivity $K$'s.}
    \end{minipage}
\end{figure}
\vspace{-10pt}

\end{document}